\documentclass[journal]{IEEEtran}

\usepackage{amsmath,amsfonts}
\usepackage{algorithmic}
\usepackage{algorithm}
\usepackage{array}
\usepackage[caption=false,font=normalsize,labelfont=sf,textfont=sf]{subfig}
\usepackage{textcomp}
\usepackage{stfloats}
\usepackage{url}
\usepackage{verbatim}
\usepackage{graphicx}
\usepackage{cite}

\usepackage[utf8]{inputenc}
\usepackage[english]{babel}
\usepackage[normalem]{ulem}
\usepackage{xcolor}
\usepackage{amssymb}
\usepackage{lineno}
\usepackage{textcomp} 
\usepackage[export]{adjustbox}
\usepackage{booktabs}
\usepackage{adjustbox}
\usepackage{nomencl}
\usepackage{multirow}
\usepackage{multicol}
\usepackage{pifont}
\usepackage{xr-hyper}
\usepackage{hyperref}

\begin{document}

\title{Extreme Scenario Selection in Day-Ahead Power Grid Operational Planning}

\author{Guillermo Terrén-Serrano, and Michael Ludkovski\thanks{Guillermo Terrén-Serrano is affiliated with the Environmental Studies Department and the Environmental Markets Labs (emLab) of the University of California Santa Barbara, Santa Barbara, CA 93106, USA. Michael Ludkovski is affiliated with the Applied Probability and Statistics Department of the University of California Santa Barbara, Santa Barbara, CA 93106, USA. Corresponding author: Guillermo Terrén-Serrano (guillermoterren@ucsb.edu).}}




\maketitle

\begin{abstract}
    We propose and analyze the application of statistical functional depth metrics for the selection of extreme scenarios in day-ahead grid planning. Our primary motivation is screening of probabilistic scenarios for realized load and renewable generation, in order to identify scenarios most relevant for operational risk mitigation. To handle the high-dimensionality of the scenarios across asset classes and intra-day periods, we employ functional measures of depth to sub-select outlying scenarios that are most likely to be the riskiest for the grid operation. We investigate a range of functional depth measures, as well as a range of operational risks, including load shedding, operational costs,  reserves shortfall and variable renewable energy curtailment. The effectiveness of the proposed screening approach is demonstrated through a case study on the realistic Texas-7k grid. 
\end{abstract}


\section{Introduction}

\IEEEPARstart{D}{aily} grid operations by Independent System Operators (ISOs) consist of carrying out day-ahead Unit Commitment (UC) to schedule generators, and then implementing throughout the day the corresponding real-time Economic Dispatch (ED) to continuously balance energy supply and demand. With the ongoing rise in power provided from solar and wind energy sources, there is major uncertainty between the respective day-ahead forecasts and realized values. This uncertainty shows up both in generation (where renewable energy sources are typically must-take during UC due to their zero marginal cost), and in load due to proliferation of behind-the-meter rooftop solar photovoltaic resources. As a result, the realized net load is highly uncertain on the day-ahead and even intra-day timescales. 

This mismatch between actuals and forecasts  
drives a wedge between the UC based on \emph{projected} renewable generation and served load, and the ED based on \emph{realized} said quantities. The associated forecast errors are a key contributor to what we call day-ahead operational risk, namely events triggering adverse outcomes during ED. These include higher than expected generation costs, deployment of operating reserves, load shedding, and Variable Renewable Energy (VRE) curtailment. From day-ahead planning perspective, it is important to identify scenarios (conditional on the knowledge of today) that carry operational risk, and to quantify their likelihood.

The first step towards this quantification entails development of statistical models for probabilistic forecasts of VRE generation and load, yielding distributions of actuals. Such probabilistic scenarios describe the full universe of \emph{what is possible} tomorrow, conditional on information available. Specifically for day-ahead UC and ED, there are probabilistic forecasting tools for day-ahead load and VRE generation that are conditional on the present weather forecast. This conditioning (rather than direct modeling of forecast errors) is essential to capture the nonlinear dependence between weather and above risks; for example, uncertainty of solar power is much lower on sunny days than on cloudy days. We use a model co-developed by the second author \cite{ludkovski2022} to produce a collection of scenarios yielding a joint probabilistic distribution of actuals across all renewable assets and zonal loads. 

In the second step, these scenarios are fed into ED simulators to obtain the resulting operational state of the grid. Aggregating ED outputs across scenarios yields the probability distribution of grid operational characteristics, such as generation costs, magnitude of reserves shortfall, load shedding and VRE curtailment. Such distributional statistics are essential for the system operator's risk management, translating the MWh-based characteristics of the scenarios into tangible operational metrics, such as dollar cost of securing generation, MWh of load shed or curtailment, etc. Ultimately, this analysis can be used to risk-adjust the UC, to secure additional capacity, or to proactively auction off additional ancillary services.

Due to the complex grid topology and transmission constraints, as well as the high-dimensional nature of the probabilistic scenarios, this second step is effectively a \emph{black box} for the ISO. Consequently, it is typically the case that there is no  rule to rank scenarios by their riskiness, rendering manual forensics or expert knowledge insufficient. 
The objective of our analysis is to provide insights regarding this black box by developing a methodology that \emph{identifies scenario subsets associated with extreme grid states}. Namely, prior to running the computationally expensive grid simulator, we aim to identify, via statistical learning methods, which scenarios are likely to lead to extreme conditions. The goal then is to \emph{pre-screen} a collection of potential actual realizations, in order to identify the extreme ones. For example, by definition any load shedding is an extreme grid state and will generally have low probability of occurrence. Thus, there might only be a handful of scenarios that entail load shed day-ahead, but the ISO would certainly like to analyze those potentialities in detail in order to best prevent/mitigate them. Rather than look for this ``needle in a haystack" across \emph{all} scenarios, our algorithm provides an efficient pre-screening that reduces the search space by an order of magnitude. As a foreshadowing, there is limited understanding on what drives VRE curtailment. Sometimes curtailment is predictable, but often it is not possible to ``foresee'' curtailment by ``staring'' at a scenario listing all the loads and VRE generation tomorrow; we provide an algorithmic solution to this challenge.

The need for statistical tools is predicated on not directly knowing which scenarios lead to operational extremality, information needed at the time of making UC decisions, but normally available only ex-post ED.
Algorithmically, we thus remove the computational bottleneck of needing hundreds of ED runs. Conceptually,  our procedure provides qualitative insights into characterizing extreme events and eliciting causal patterns in grid behavior. 

In a complementary use case, advanced UC methodologies are supplanting  point forecasts  with a (small) collection of alternative scenarios via a Stochastic Programming (SP) paradigm. For effective risk management, one goal of SP is to select a small set of \emph{extreme scenarios} such that the UC algorithm can anticipatively resolve operational risk; see for example security-constrained UC with stochastic wind generation in \cite{zhu2020}. This raises the same challenge: given a large collection (1000+) of potential scenarios, identify a subset ($ \ll \!100$) of those that will lead to operational extremality. Scenario selection also arises during day-ahead grid \emph{stress testing}.  In present, stress testing is often done manually or targets specific events like hurricanes \cite{satkauskas2022}, but we anticipate the switch to probabilistic scenarios whereby we have a large collection of scenarios and need to identify which ones are best to carry out the ED stress tests. 

For all of the outlined applications, the abstract task can be viewed as translating probabilistic load/generation scenarios into extremality scores. Thus, we wish to either \emph{rank} scenarios in terms of their operational extremality, or to classify them into \emph{risky/non-risky}, according to a given output of interest (such as magnitude of load shedding, or total generation cost). This  task is highly nontrivial due to the very high dimension of the probabilistic scenarios. Indeed, each scenario describes the joint behavior of $A$ assets across $T=24$ hours, yielding $T \times A \gg 10^3$ total data points for realistic systems with $A \gg 10^2$ generators. Therefore, traditional or manual methods to do forensics in order to understand the dependence of outputs on inputs, or risk classification are infeasible. 

To this end, we propose to employ  statistical \emph{functional data analysis}, especially functional data depth, techniques. The key idea is to view scenarios as \emph{functional} objects, encompassing their temporal structure that is viewed holistically during the day-ahead planning stage. Functional depth is a common tool for statistical classification problems of functional data \cite{cuevas2007,hubert2017,sguera2014,zhang2021}. To our knowledge, we are the first to use functional depth for grid operational planning. Our novel proposal can be contrasted with existing strategies for selecting extreme scenarios, such as clustering techniques \cite{luo2020} or principal component analysis \cite{farjad2019}. The above approaches are infeasible in our setting due to the ultra-high-dimensionality of the scenarios in question. 

To sum up, our contribution is to propose a new link between scenario selection for grid operational planning and statistical functional analysis. Specifically, we apply statistical depth measures to rank and pre-screen scenarios for operational extremality. Our analysis demonstrates the effectiveness of this idea and presents an extensive case study for a realistic grid mimicking the Electric Reliability Council Of Texas (ERCOT) serving region within the U.S. 

The rest of the article is organized as follows. Section \ref{sec:methods} defines the operational and statistical measures of extremality. Section \ref{sec:Texas} describes our experimental setup. The results are in Section \ref{sec:results} where we go through four different case studies, covering detection of high generation costs, high reserves shortfalls, load shedding and VRE curtailment. We discuss our take-aways in Section \ref{sec:discuss}; Section \ref{sec:conclude} presents outlook for future analysis.

\section{Methods}\label{sec:methods}

\subsection{Extremality in Operational Planning}\label{sec:extremality_def}

A scenario for day-ahead load and VRE generation is \emph{extreme} if the respective realization poses a risk for the regulation of a power grid by the ISO. For example, scenarios where load is higher than expected might be extreme in terms of potential load shedding, while scenarios where generated renewable energy is much higher than forecast might be extreme in terms of potential VRE curtailment.  Broadly speaking, extremality is caused by forecasting \emph{errors}. Discrepancies between forecasts and actuals require adjustments during ED, with the effect of the ISO having to either commit additional units, or reduce the energy generated from the committed units to rebalance the energy demand and supply.
Pushed to the extreme, forecast errors can trigger load shedding or VRE curtailment. But even without those drastic outcomes, forecast errors inevitably lead to additional expenditures and inefficiencies. 

Both UC and  ED are carried out on a holistic basis taking into account all the grid transmission and congestion constraints, as well as its temporal structure (such as multi-hour ramping of thermal units). Thus,  hour-by-hour mismatches between forecasts and actuals are not sufficient to identify extreme scenarios. Consequently, risk managers must adopt a similar holistic perspective, treating scenarios as \emph{curves} indexed in time (a time-series perspective is also limiting since the UC considers the whole daily profile at once rather than hour-by-hour).

Accordingly, we adopt the functional analysis perspective, viewing our data as a  collection $\mathcal{F} = \{\hat{f}^{\mathcal{R}}_{i,k}(\cdot) \in \mathcal{C} \left( \mathcal{I} \right), i = 1, \dots, N\}$ of $i=1,\ldots,N$ scenarios for operational day $k$, indexed by the respective resource class $\mathcal{R} = \{\mathcal{S}, \mathcal{W}, \mathcal{L}\}$ representing Solar, Wind and Load assets. Above the time index set is $t \in \mathcal{I} = \left[1, T\right]$ where $T = 24$ are the hours \cite{ludkovski2022}. The (point) forecast can be thought of as the average scenario, $\mu^{\mathcal{R}}_k(t)= Ave_i( \hat{f}^{\mathcal{R}}_{i,k}(t))$, so that  $\hat{f}^{\mathcal{R}}_{i,k}(t)-\mu^{\mathcal{R}}_{i,k}(t)$ is the forecast error, i.e.~the mismatch between the prediction and realized actuals, for scenario~$i$.

\subsection{Statistical Extremality}\label{sec:functional_depths}

The intuitive characterization of operational extremality is scenarios that are ``outlying'', i.e.~are far from their bulk and the forecast (interpreted as the ``central'' scenario). Hence, we aim to quantify the notion of statistical outlyingness for  day-ahead scenarios (viewed as functional objects), equating  it with operational risk. One challenge is that \emph{a priori} it is not clear how to define outlyingness, especially since there are multiple facets of operational risk, for example the risk of high generation costs, or the risk of renewable curtailment. 

To impose center-outward ordering on multivariate objects, we generalize the univariate constructs of empirical quantiles and outliers by adopting the concept of \emph{functional depth} $F(f; {\cal F})$ of a given functional object $f$ relative to a collection ${\cal F}$ of functions. The positive-valued depth function $F$ assigns \emph{depth scores} in the range $[0,1]$, where a depth of $1$ corresponds to the least outlying, i.e.~a functional ``centroid'', of the collection ${\cal F}$. Conversely, outliers have depth scores close to zero. Thus, the ``more central'' an object is with respect to ${\cal F}$, the higher its functional depth score. 

Multiple approaches exist to construct notions of functional depth. Some of them originate in functional data analysis \cite{wang2016}, while others are concerned with multivariate statistics. A key motivating application is outlier detection. To offer a brief taxonomy of depth functions, we distinguish several sub-classes. \emph{Rank-based} depth functions utilize the marginal rank 
$$R_i(t) = \sum_j 1_{\{\hat{f}_j(t) < \hat{f}_i(t)\} } \in \{1,\ldots,N\}$$
of $\hat{f}_i(t)$ within the set $\{ \hat{f}_j(t): j=1,\ldots,N\}$.  In contrast \emph{distance-based} depth functions employ a metric $\| \hat{f}_i - \hat{f}_j \|$. Rank-based metrics are less sensitive to relative magnitudes and more geared to sorting ${\cal F}$, while distance-based metrics home in on magnitude outliers. From a different perspective, depth functions can be integrated or not. Integrated depth is based on defining a univariate notion of outlyingness for each marginal value $\hat{f}_i(t), t \in {\cal I}$ and then aggregating (through a sum or a maximum) the resulting ``curve of outlyingness'' across $t$'s. Non-integrated notions treat the entire $\hat{f}_i$ holistically.  Finally, depth functions can target \emph{magnitude} or \emph{shape} outliers. A scenario for day-ahead load might be unusual in magnitude (very high load relative to forecast), or unusual in its shape (e.g. achieving its peak/crest and trough at hours other than forecast). Magnitude outliers can be thought of as being extreme at some part of the time-domain. Shape outliers have an unusual overall functional shape (e.g.~extreme ramp, i.e.~gradient in time), even if they are not extreme at any given hour. 

Below we recall 8 different notions of functional depth for defining the concept of extremality in stochastic day-ahead operational planning of power grids. We cite their original proposal in statistical literature, none them having been used in grid applications.  Throughout, we provide definitions for the depth $F_i$ of scenario $\hat{f}_i$ relative to the scenario database ${\cal F}$. 

\subsubsection{Integrated Depth (ID)}\label{sec:ID}
is rank-based and averages the univariate ranks $R_i(t)$ across the hours \cite{fraiman2001}:
\begin{equation}
    ID_i := 1 - \frac{1}{T}\int_{\mathcal{I}} \left| \frac{1}{2} - 
    \frac{R_i(t)}{N}\right| dt.
    \label{eq:ID}
\end{equation}

\subsubsection{Modified Band Depth (MBD)}\label{sec:MBD}
is also an integrated rank-based metric. MBD averages the product $(R_i(t)-1)(N-R_i(t))$, which is maximized at the median, across time:
\begin{equation}
    MBD_i := \frac{1}{N (N + 1)} \int_{\mathcal{I}} \frac{ \left( R_i (t) - 1 \right) \left( N - R_i (t) \right) }{N - T - 1} dt.
\end{equation}
The MBD underlies the commonly used functional boxplot for identifying magnitude outliers.

\subsubsection{Extremal Depth (EXD)}\label{sec:ED}
is another rank-based measure originally due to \cite{narisetty2016}. It is based on the pointwise depth
\begin{equation}
    d_i (t) := 1 - \frac{\left| 2R_i(t) - N - 1\right|}{N}.
    \label{eq:EXD}
\end{equation}
Rather than integrating $d_i(\cdot)$, EXD constructs the depth Cumulative Density Function (d-CDF)
\begin{equation}
    \Phi_i(r) := \frac{1}{T} \int_{ \mathcal{I}} \mathbb{I} \left( d_i (t) \leq r \right) dt, \qquad r \in (0,1)
\end{equation}
and then ranks $\hat{f}_i$ according to the left-tail stochastic ordering of $\Phi_i$'s. Thus, the $EXD_i$ of a function $\hat{f}_i$ is defined as  
\begin{equation}
    EXD_i := \frac{1}{N} \sum_{j=1}^N \mathbb{I}( {\Phi}_{i} \succcurlyeq {\Phi}_{j}) ,
\end{equation}
where the stochastic order $\Phi_{j} \succcurlyeq \Phi_{i}$ intuitively means that the frequency (in $t$) of $\hat{f}_i$ achieving an extreme rank is higher than that for $\hat{f}_j$.

\subsubsection{Extreme Rank Length Depth (ERLD)}\label{sec:ERLD}
is based on the left-tail stochastic ordering $d_i \succcurlyeq d_j$ of the univariate depth curves $d_i(\cdot)$ from \eqref{eq:EXD} in terms of their sorted depth values,
\begin{equation}\label{eq:ERLD}
    ERD_i := \frac{1}{N} \sum_{j=1}^N \mathbb{I} \left( d_i \succcurlyeq d_j \right).
\end{equation}

\subsubsection{L-Infinity Depth (LID)}\label{sec:LD}

We next move on to distance-based depth functions. 
The $L_\infty$ Depth \cite{zuo2000} is based on the average distance between $\hat{f}_i$ and other $\hat{f}_j$'s, using the Chebyshev distance $\lVert \hat{f}_j - \hat{f}_i \rVert_\infty = \max_t |\hat{f}_j(t) - \hat{f}_i(t) |$,
\begin{align}\label{:LID}
    LID_i & := \frac{1}{1 + \frac{1}{N}\sum_{j = 1}^N \lVert \hat{f}_j - \hat{f}_i \rVert_\infty}.
\end{align}
Other choices of functional distance can be selected and substituted into $L_i := \frac{1}{1+ \mathbb{E}_F[ |f-F| ]}$ where the expected value is interpreted as the empirical average across elements in ${\cal F}$.

\subsubsection{h-Mode Depth (HMD)}
 \cite{cuevas2007} is also distance-based and evaluates the  ${\cal F}$$-$``density'' in the neighborhood of $\hat{f}_i$ by computing the pairwise distances $\| \hat{f}_i - \hat{f}_j\|$ and then using a kernel density estimator
\begin{equation}
    HMD_i := \frac{1}{N} \sum_{j = 1}^N \mathcal{K} \left( \lVert \hat{f}_i  - \hat{f}_j \rVert \right).
    \label{eq:HD}
\end{equation}
We use the $L_2$-norm $\| \hat{f}_i - \hat{f}_j\|^2_2 = \sum_t |\hat{f}_i(t)-\hat{f}_j(t)|^2$ and the Gaussian kernel $\mathcal{K} (z) = (\sqrt{2\pi} h)^{-1} \exp \left(- z^2/(2 h^2)\right).$ The bandwidth $h$ is automatically tuned to the 15\% quantile of the pairwise distances in ${\cal F}$.

\subsubsection{Directional Quantile (DQ)}\label{sec:DQ}
 \cite{dai2020} is a distance-based approximation to ERLD in Eq.~\eqref{eq:ERLD},
\begin{multline}    
    DQ_i := \max_{t \in {\cal I}} \Bigg\{ \mathbb{I}( \hat{f}_i(t) \geq \mu (t) )\cdot \frac{\hat{f}_i(t) - \mu (t)}{\left| Q_{97.5}(t) - \mu (t) \right|} \\
    + \mathbb{I} ( \hat{f}_i(t) < \mu(t) ) \cdot  \frac{\hat{f}_i(t) - \mu (t) }{\left| Q_{2.5} (t) - \mu (t) \right|} \Bigg\}, 
\end{multline}      
where $Q_{97.5} (t)$ and $Q_{2.5} (t)$ are the respective pointwise quantiles of $\{\hat{f}_j(t)\}$, and $\mu (t)$ is the pointwise mean of $\{\hat{f}_j(t)\}$.

\subsubsection{Random Tukey Depth (RTD)}\label{sec:TD}
\cite{cuesta2008} is our final choice. Unlike all other approaches above, RTD yields a \emph{random} measure of functional depth. Specifically, RTD is based on the univariate ranks of a projection $\langle \hat{f}_i, v_k \rangle \equiv \sum_t \hat{f}_i(t)v_k(t)$ of $\hat{f}$'s along random directions $v_k$,
\begin{multline}\label{eq:rtd}
    RTD_{i} := \\ \hspace*{-10pt}
    \min_{k=1,\ldots,K} \Bigl\{ 1 - \frac{|2 \sum_j \mathbb{I}( \langle \hat{f}_i, v_k \rangle \le \langle \hat{f}_j, v_k \rangle ) - N - 1|}{N} \Bigr\}.
\end{multline}
Thus, the sorting is based on comparing $\langle \hat{f}_i, v_k \rangle$ across ${\cal F}$, compare to Eq.~\eqref{eq:EXD}; the minimum in Eq.~\eqref{eq:rtd} is inspired by the original Tukey Depth which takes the minimum depth across \emph{all} $v_k \in \mathbb{R}^T$, understood as minimizing the depth of $\hat{f}_i$ relative to ${\cal F}$ among all half-spaces in $\mathbb{R}^T$. We use $K = 50$ random projections and sample the projection vectors $v_k$ from the i.i.d. standard Gaussian distribution.

\subsection{Extreme Scenarios Selection}

To predict operational extremality, we apply the above notions of statistical extremality on different facets of the scenarios. Specifically, we consider \emph{load} $ \hat{f}_i^{\mathcal{L}}$, \emph{solar} generation $ \hat{f}_i^{\mathcal{R}}$, \emph{wind} generation $ \hat{f}_i^{\mathcal{W}}$, renewable (\emph{VRE}) generation $\hat{f}_i^{\mathcal{G}}(t) := \hat{f}_i^{\mathcal{S}}(t) + \hat{f}_i^{\mathcal{W}}(t)$ and \emph{net load}, 
$$\hat{f}_i^{\mathcal{N}}(t) := \hat{f}_i^{\mathcal{L}}(t) - \hat{f}_i^{\mathcal{G}}(t) = \hat{f}_i^{\mathcal{L}}(t) - \hat{f}_i^{\mathcal{S}}(t) -\hat{f}_i^{\mathcal{W}}(t).$$
The above quantities can be defined at the aggregate grid- or zonal-level. We explore both alternatives, starting with the grid-level first and then adding zonal-level information to enhance detection accuracy for load shedding and VRE curtailment.

The functional depth notions introduced in Section~\ref{sec:functional_depths} impose center-outward ranking. Therefore, the resulting depth metrics select both ``top'' and ``bottom'' scenarios, interpreted as scenarios with the highest/lowest  MWh magnitude. To focus on scenarios with highest net load and/or lowest VRE generation, we add a pre-screening step based on Area Under the Curve (AUC). The AUC is simply the integral of the ${\cal R}$-facet of a scenario across the 24 hours, $a^{\mathcal{R}}_i := \sum_{t = 1}^T \hat{f}_i^{\mathcal{R}}(t)$. We primarily consider AUC by net load, $a^{\cal N}$. 

\section{Experimental Setup}\label{sec:Texas}

\begin{figure}[!htb]
    \centering
    \includegraphics[scale = .35, trim = {0cm, 0.5cm, 0cm, 0.2cm}, clip]{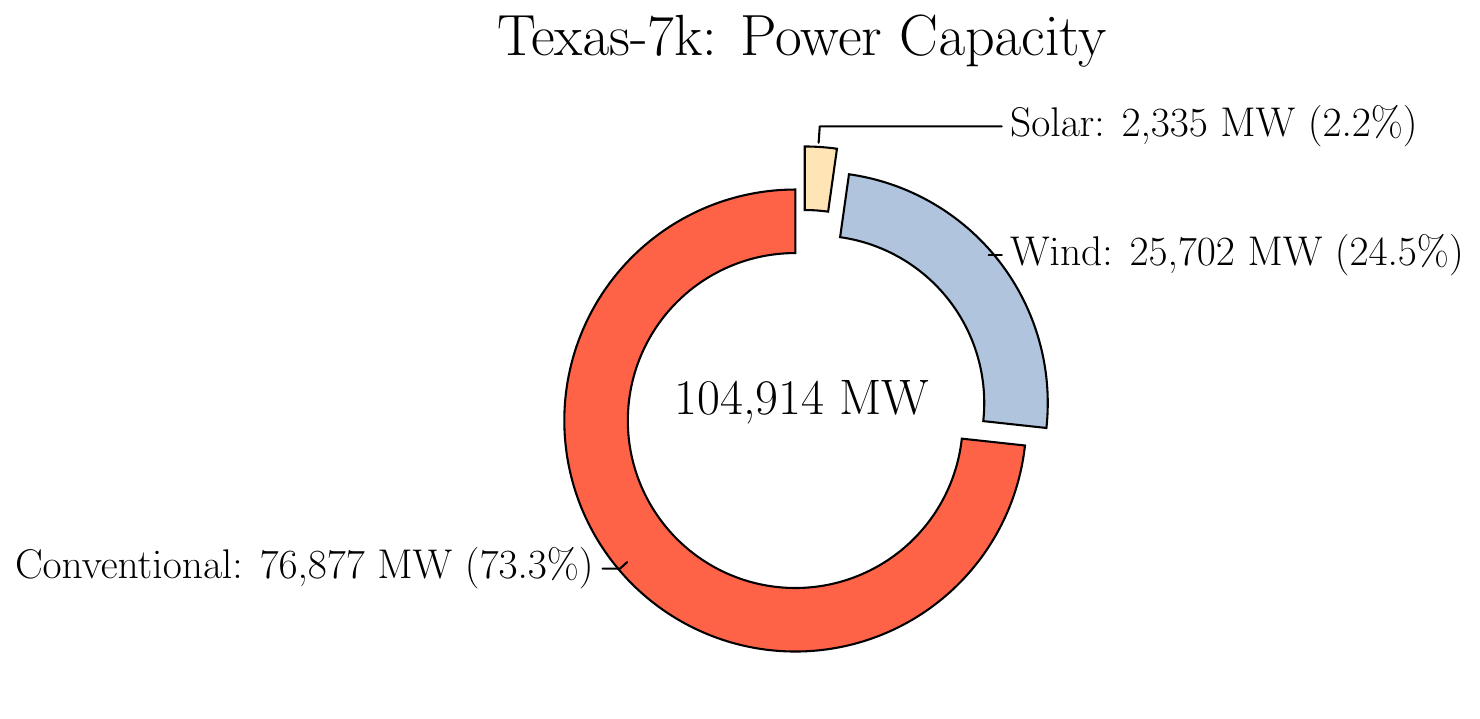}
    \caption{Fuel mix in the Texas-7k power grid. Total power capacity is 104.9~GW, with 26.73\% from VRE generators. 
    }
    \label{fig:load_shedding_and_energy_curtailment}
\end{figure}

For our experimental testbed we use a synthetic representation of the ERCOT grid. This Texas-7k grid covers the geographic area operated by ERCOT, representing the majority of the U.S. state of Texas \cite{birchfield2018}. It has 6,716 buses and 185 VRE generators: 36 solar and 149 wind, see fuel mix and nameplate power capacity in Fig.~\ref{fig:load_shedding_and_energy_curtailment}. The network topology includes transmission and distribution lines with transformers. cf.~Fig.~\ref{sm:topology}. In analogy with ERCOT, Texas-7k has eight balancing zones: North, North Central, East, Coast, South Central, South, West, and Far West which are used to model loads. 

\subsection{Probabilistic Scenarios}

The day-ahead forecasting scenarios we work with are conditional to the forecasts of energy demand and asset-level generation. The original dataset of forecasts and actuals was developed by NREL using re-analyzed numerical weather predictions from 2017 and 2018. The re-analyzed ensembles are from the ERA5 hourly data maintained by the European Centre for Medium-Range Weather Forecasts (ECMWF). The solar irradiance and wind speeds forecasts had a 10~km spatial resolution and a frequency of 5~minutes and were interpolated to obtain energy generation in MWh using an asset-specific physical model. The forecasting scenarios were specific to wind and solar assets (i.e.~power plants) and load-balancing zones. 

Probabilistic scenarios were generated using the \emph{clnSim} framework \cite{ludkovski2022} and provide full uncertainty quantification around the above forecasts. This framework calibrates the day-ahead scenarios to the specific calendar date and employs hierarchical clustering to efficiently and robustly estimate the full covariance matrix across hundreds of assets and 24 hours of the day. Covariance estimation employs a structured Gaussian copula after de-trending the joint dependence between forecasts and actuals, and taking into account the multiple layers of seasonality. \emph{clnSim} generates arbitrary number of scenarios for a given set of assets; in our case we generated 1000 scenarios at hourly resolution across the 8 ERCOT zones and the 185 VRE assets.

\subsection{Simulating UC/ED Operations}

To obtain grid operational metrics, scenarios are inputted into UC and ED solvers. Specifically, we use the \textit{Vatic} Python library \cite{grzadkowski2021} that simulates the day-ahead market and resulting daily grid operations. \textit{Vatic} uses a Mixed-Integer Linear Programming (MILP) formulation \cite{Egret2020, knueven2020} to solve the UC problem and relies on \textit{Pyomo} \cite{bynum2021} open-source optimization engine. We employ \textit{Gurobi} as the low-level, licensed optimization software \cite{gurobi2023}. The ED is solved with a 2-hour horizon and a 20\% operating reserve margin. The day-ahead UC has time resolution of next 24 hours. The initial  states of the thermal generators were generated by simulating the operation of the entire 2018 with actual data. Note that this means identical initial conditions for all 1000 scenarios on a given day.

After solving for UC and ED, \textit{Vatic} generates hourly reports for the variables of interest, namely variable generation costs $G_i(t)$ [\$] (i.e.,~costs involved in ED), reserves shortfall (RS) $R_i(t)$ [MWh], load shedding (LS) $L_i(t)$ [MWh], and VRE curtailment (VC) $C_i(t)$ [MWh]; it also reports the congestion in transmission lines and the capacity factors for the committed generators (see Fig.~\ref{sm:vatic_simulations}). RS refers to the one-sided deviation (higher load or under-generation) of actuals relative to forecast and represents deployment of operating reserves during ED. Thus, RS closely relates to the above operating reserve margin (fixed at 20\%) and relates to the reliability risk were the generating capacity to be exhausted. We generally add up the hourly metrics across 24 hours to obtain a single daily metric of grid performance, e.g.~$L_i := \sum_{t=1}^{24} L_i(t)$. 


We define operational extremality \emph{relatively} for a given day. The output statistics $G_i$ and $R_i$ effectively take values on a continuum, and we seek the top 5\% of scenarios according to that facet, reflecting the typical risk threshold.
$L_i$ and $C_i$ are generally zero, and we consider a scenario extreme if it involves some load shedding or VRE curtailment, see Sections \ref{sec:load-shed}-\ref{sec:curtailment} below.
 

Since each UC/ED simulation takes close to a minute, in our case studies we restricted grid daily simulated to 25 test days randomly and uniformly chosen from the 2018 calendar year (Jan~2 and 20, Feb~13, 14 and 26, Mar~14, Apr~1, 9 and 24, May~10, 27, Jun~4, 30, Jul~22, 24, Aug~8, 18, Sep~4, 14, Oct~4, 17, Nov~2, 13, and Dec~1, 27). The statistics related to the number of load shedding and VRE curtailment scenarios for each of those test days are in Fig.~\ref{sm:load_shedding_and_energy_curtailment}. We emphasize that our ED simulations are based on a non-security constrained UC solution, and are meant to explore potential risks to the grid across plausible day-ahead scenarios for load and VRE generation. Consequently, extreme grid conditions are encountered relatively frequently. For example, a non-negligible number of days entail the potential for load shedding. VRE curtailment is even more frequent: among the 25 chosen days and 1000 scenarios per day, there are an average of 149.48 hourly load shedding events totaling 245.04~GWh, and 241.48 hourly VRE curtailment events totaling 87.23~GWh. 
Load shedding events are more frequent and larger in magnitude during the summer, reflecting that ERCOT peak energy demand is driven by the summer high temperatures. VRE curtailment is most likely when the grid relies on wind resources in the fall and winter months. 

\begin{figure*}[!b]
    \centering
    \includegraphics[scale = .5, trim = {1.25cm, 0cm, .5cm, 0cm}, clip]{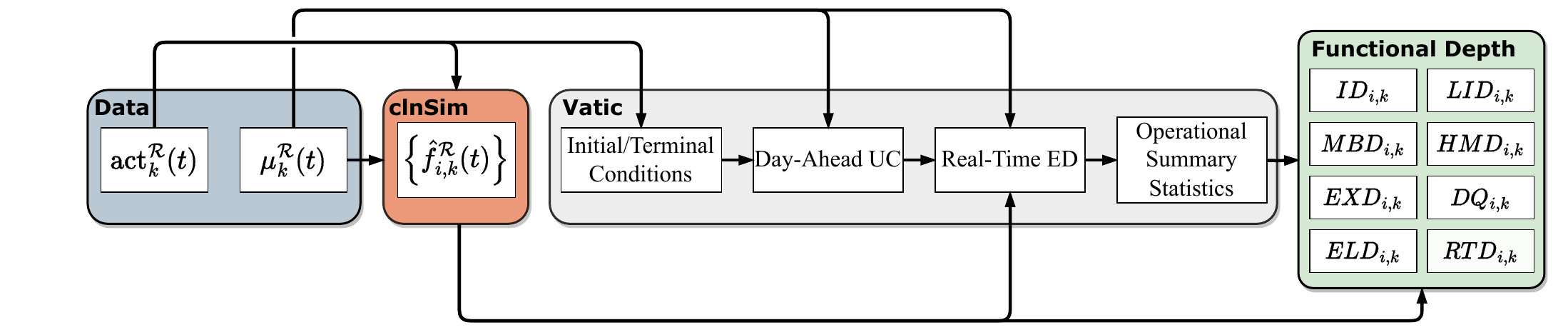}
    \caption{Case study workflow to efficiently screen \emph{clnSim} scenarios necessary for day-ahead power grids operational planning in \emph{Vatic}.}
    \label{fig:workflow}
\end{figure*}

\begin{figure*}[!b]
    \centering
    \begin{tabular}{cccc}
    \includegraphics[scale = 0.245, trim = {0cm, 0cm, 0cm, 0cm}, clip]{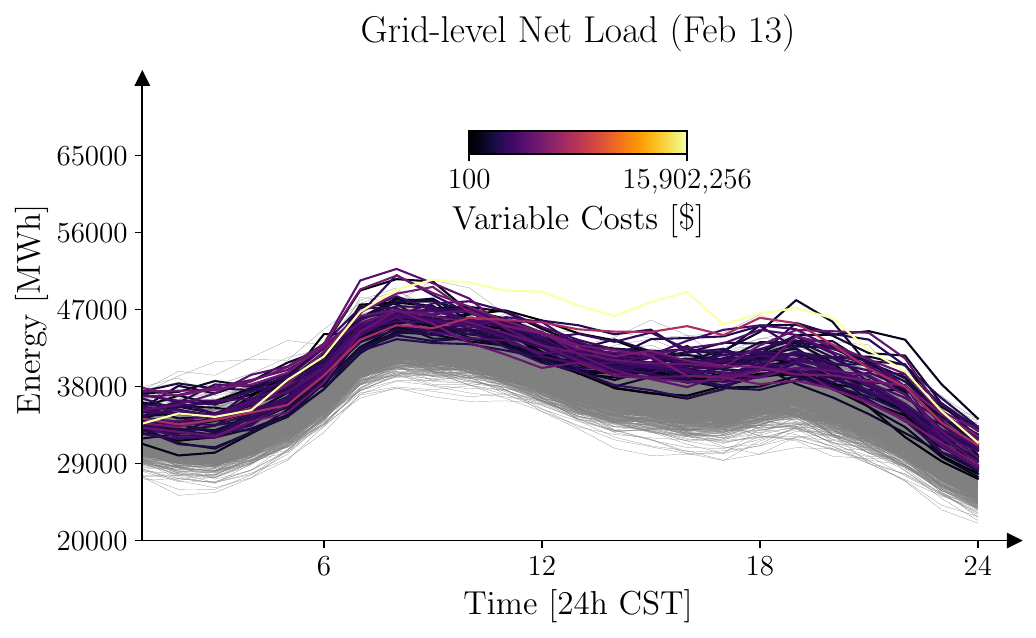} & 
    \includegraphics[scale = 0.245, trim = {.9cm, 0cm, 0cm, 0cm}, clip]{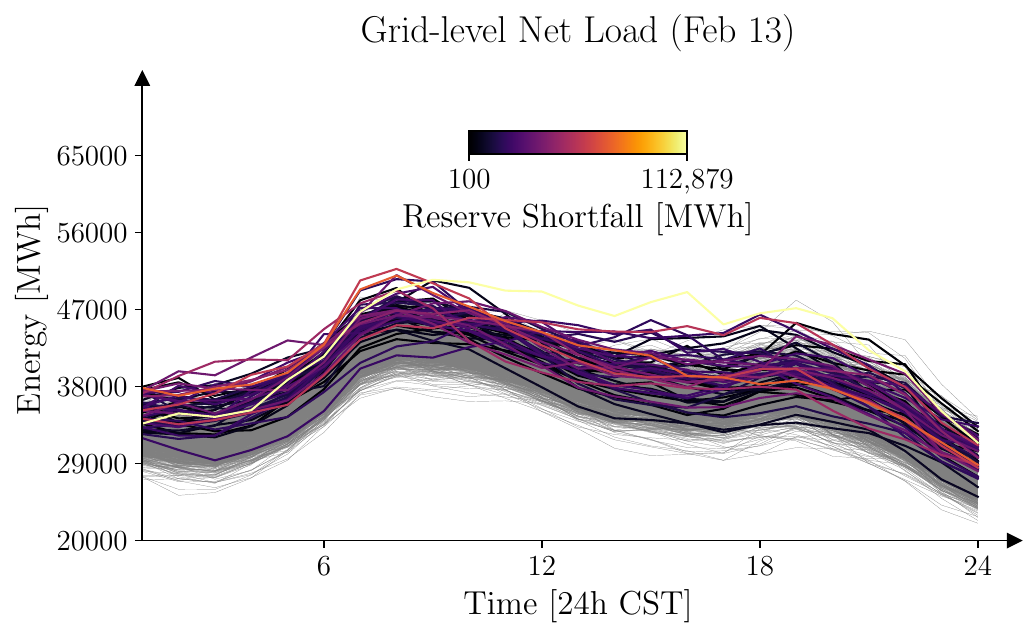} & 
    \includegraphics[scale = 0.245, trim = {.9cm, 0cm, 0cm, 0cm}, clip]{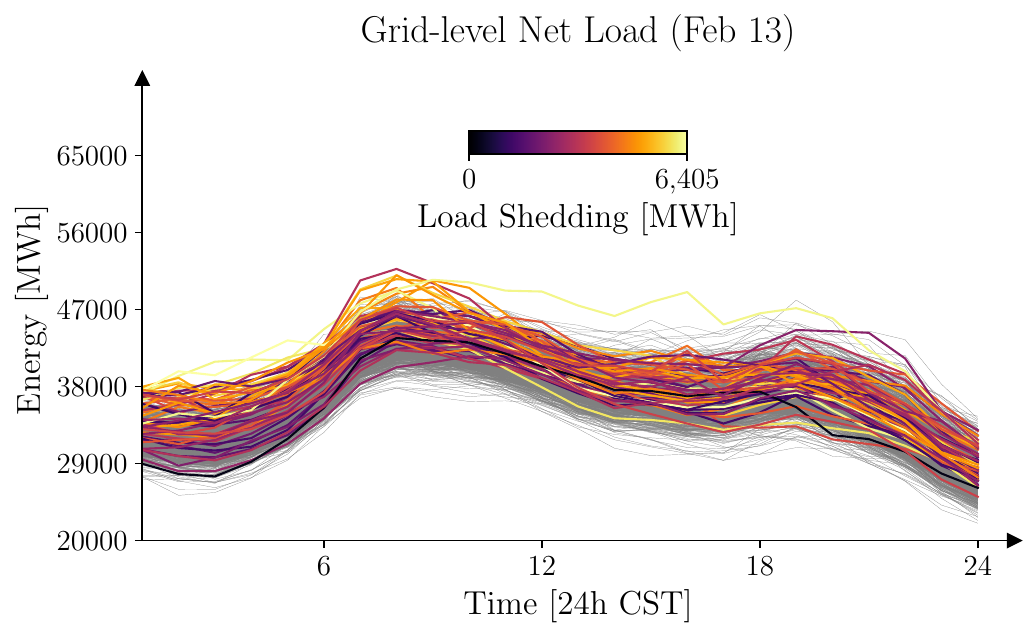} & 
    \includegraphics[scale = 0.245, trim = {.9cm, 0cm, 0cm, 0cm}, clip]{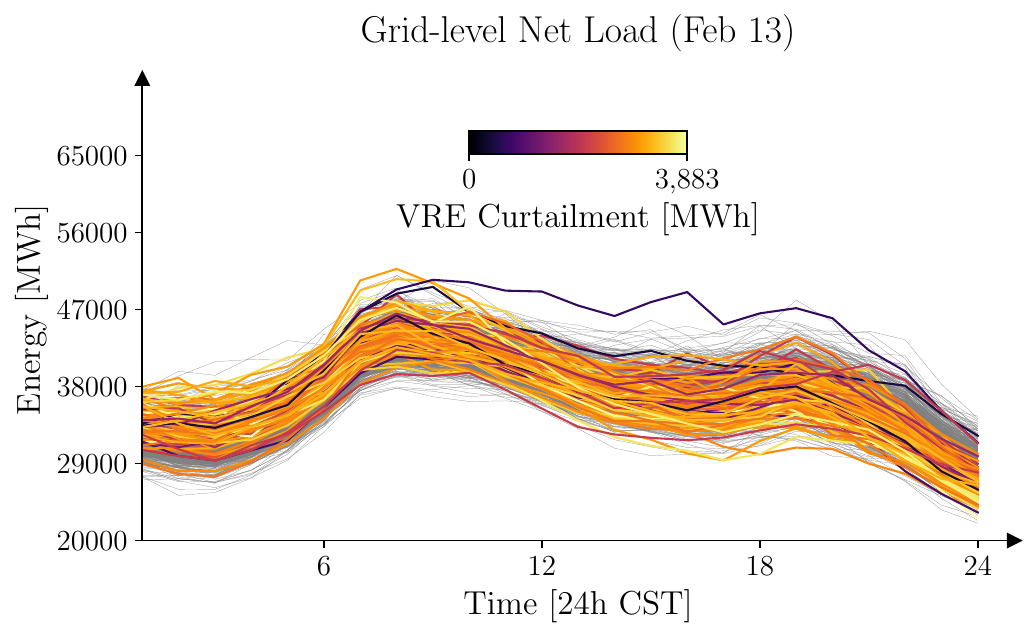} \\
    $G(t)$ & RS $R(t)$ & LS $L(t)$ & VC $C(t)$ 
\end{tabular}
    \caption{Four facets of operational extremality against aggregated hourly net load $f^{\cal N}_i(t)$ on Feb~13, 2018. We highlight the top $m=50$ scenarios in terms of each facet: the more extreme a scenario, the brighter its color.}
    \label{fig:agg_net_load_extreme}
\end{figure*}


\subsection{Extremality Detection}

Fig.~\ref{fig:workflow} summarizes our overall workflow. In the absence of any statistical selection procedure, each probabilistic scenario must be fed through the grid UC/ED simulator to assess its operational risk. Our goal is to reduce this computational burden by first computing a suitable functional depth metric for scenarios in order to pre-screen them. 
Let ${\cal E}_k$ denote the set of operationally extreme scenarios on day $k$.
We look to construct a set ${\cal O}^{\cal D}_k$ that best predicts ${\cal E}_k$ via a given functional depth ${\cal D}$.  Let $m = |{\cal E}_k|$ and $n=| {\cal O}^{\cal D}_k|$; to allow a margin of mis-identification, we take $n > m$, typically $n \simeq 1.5 m$.
Given $m$ and $n$, our accuracy metric is
\begin{equation}
    P^{\cal D}_k := \frac{ | {\cal E}_k \bigcap {\cal O}^{\cal D}_k| }{| {\cal E}_k|} \in [0,1],
\end{equation}
where the numerator counts the number of selected scenarios that are operationally extreme and the denominator is the number of extreme scenarios. Since $n \gg m$, we aim to achieve close to 100\% accuracy $P_k \simeq 1$, i.e.~identify nearly all extreme scenarios among those selected. Alternatively, for load shedding and VRE curtailment, we weigh accuracy by the magnitude (in GWh) of the risk, 
\begin{equation}
    P^{\cal D}_k :=  \frac{  \sum_{j \in {\cal O}_k} \sum_{t = 1}^T X_{j,k} (t)}{ \sum_{i \in {\cal E}_k} \sum_{t = 1}^T X_{i,k} (t)} \in [0, 1],
\end{equation}
where $X_{k}(t) \in \{L_k(t), C_k(t)\}$ represents a given operational facet of the scenarios. 


\section{Selecting Scenarios for Different Extremality Criteria}\label{sec:results}

We proceed to apply our methodology to detect extreme day-ahead scenarios in the described Texas-7k grid, targeting the four aforementioned operational statistics: (i) generation costs; (ii) reserves shortfall; (iii) load-shedding and (iv) VRE curtailment. These are arranged in increasing order of selection difficulty and are addressed in turn in the following subsections. The functional depth metrics from  Section~\ref{sec:functional_depths} are  evaluated to find which statistical metric best aligns with operational extremality.

To illustrate operational extremality, Fig.~\ref{fig:agg_net_load_extreme} visualizes the $m = 50$ most risky scenarios across the above four facets on one of our test days. Since grid risk is intuitively tied to system Net Load (NL), we show the $N = 1000$ scenarios in terms of their hourly aggregated NL. Fig.~\ref{fig:agg_net_load_extreme} demonstrates that there is a clear link between NL and operational costs, as well as reserves shortfall. However, there is little dependence between NL and load shedding or VRE curtailment. These varying relationships between net demand and operational risk is the motivation to consider in the sequel statistical tools to predict extremality. 


Our first metric is generation costs $G_i$. Highest variable costs occur in summer when NL is largest, though high NL can also occur in winter during extreme cold weather. For detecting extreme $G_i$, 
we observe a very strong correlation (median $R^2$ of 0.997 for the daily linear regression) between NL $f^{\cal N}_i$ and $G_i$. Thus, the conventional approach of selecting scenarios based on daily net load AUC is sufficient. 
Indeed, taking ${\cal O}_k$ to be the set of 50 scenarios with the highest daily total NL, and ${\cal E}_k$ to be the set of 50 scenarios with the highest costs, we have $Ave_k( | {\cal E}_k \bigcap {\cal O}_k|) = 46.84$ yielding average accuracy of over 93\%, and if we expand to top-75 scenarios based on NL, then on average we can identify $Ave_k(| {\cal E}_k \bigcap {\cal O}^{(75)}_k|) = 49.96$ out of top 50.
The take-away is that ED simulation is quite transparent when it comes to variable generation costs, and the latter can be easily predicted by the univariate AUC statistic. Consequently, there is no need (or benefit) from applying further statistical tools for this task. 

\subsection{Magnitude of Reserves Shortfall}\label{sec:shortfall}

\begin{figure}[!htb]
    \centering
    \includegraphics[scale = 0.33, trim = {0cm, 0cm, 25cm, 0cm}, clip]{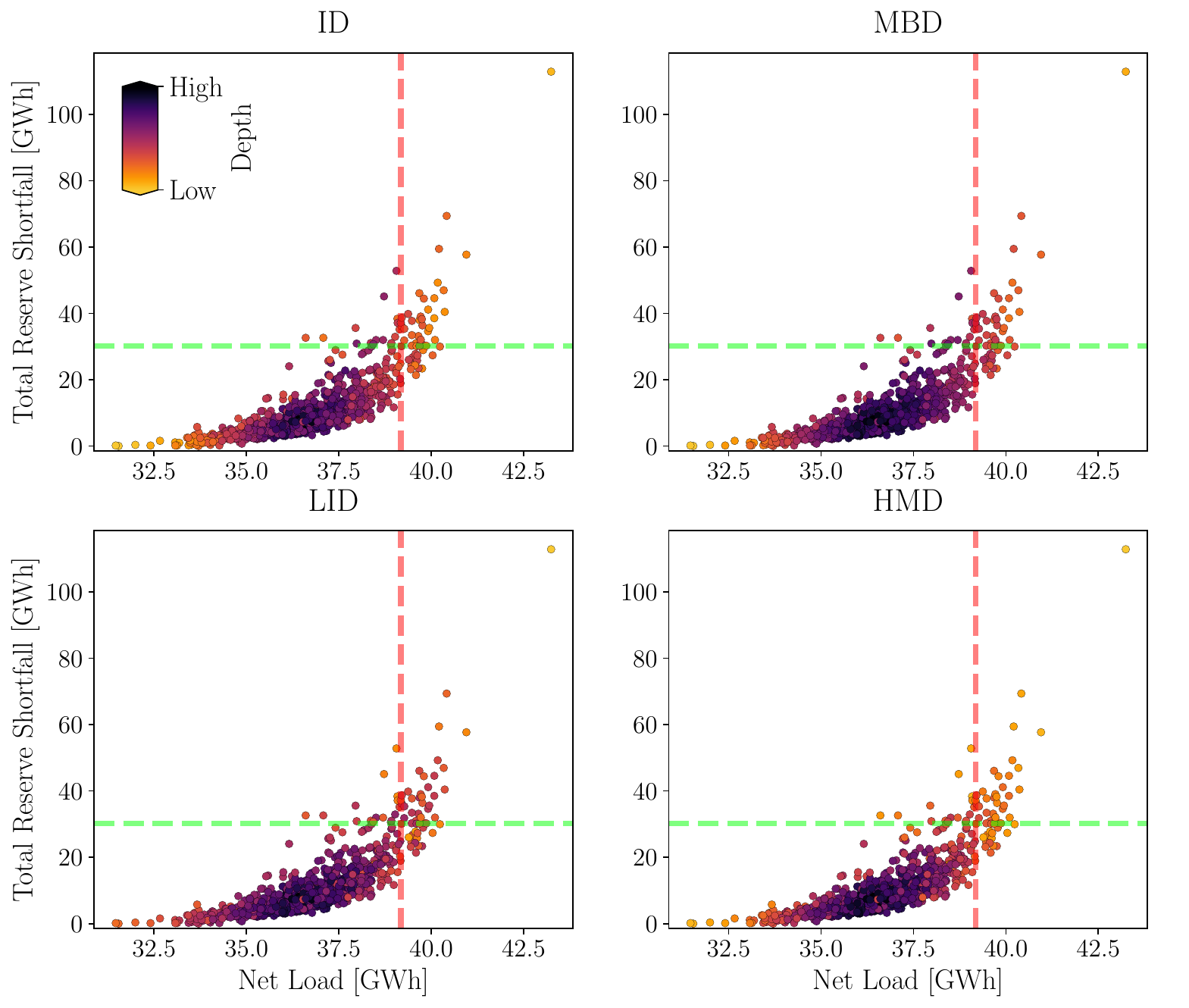}
    \caption{Detecting reserves shortfall via 4 different functional depth metrics on Feb~13, 2018. Aggregated daily NL on the $x$-axis vs.~daily RS on the $y$-axis. Scenarios are color coded according to the respective depth metric: the brighter the color gradient, the less deep the scenario. Green horizontal (vertical red) line at 30.17~GWh (resp.~39.18 GWh) shows the threshold for the top 50 highest RS (resp.~top 50 highest NL).}
    \label{fig:scatter_net_vs_shortfall}
\end{figure}

We move on to selecting scenarios that have the highest Reserve Shortfall (RS). 
Our objective is to utilize functional depth to identify the most likely scenarios to be in the top $5\%$ (i.e.,~$m=50$ out 1000 total) for RS.
Fig.~\ref{fig:scatter_net_vs_shortfall} plots RS vs.~NL and the various functional depth metrics, so that the prediction task is to identify the ``top'' (in terms of the $y$-axis) scenarios. Though NL exhibits a substantial positive correlation with RS, selecting the top $n$ ``right-most'' (i.e.~highest NL) scenarios is not sufficient to accurately identify $R_i$-extreme scenarios. Indeed, there are many scenarios in the bottom-right corner with high NL but low RS. 

Ranking by functional depth remedies this by incorporating statistical outlyingness. For example, we observe that MBD is effective in selecting the top-right scenarios (true positives) while filtering out the bottom-right (false positives) ones. Fig.~\ref{fig:scatter_net_vs_shortfall} highlights the scenarios in the top 5\% by RS (horizontal green line) vis-a-vis the scenarios in the top 5\% by NL (vertical red line), showing that ranking by functional depth improves the respective proportion of false positive (bottom-right quadrant) and false negatives (top-left quadrant). 


We devise the following two-stage procedure to select $75 = 1.5 m$ scenarios. Matching the standard top-down approach, we use only aggregated grid-level generation and load as predictors.  In stage 1, the $n_1$ scenarios ($n_1 < N$) with the greatest AUC by NL are selected from the whole set ${\cal F}$, while the remaining $N-n_1$ are screened out. In stage 2, $n_2$ scenarios ($n_2 < n_1$) are selected among these based on the lowest depth of a given functional metric. In all, we combine ranking by functional depth (see Fig.~\ref{sm:scatter_rank_vs_shedding}) with filtering out the lowest-AUC scenarios, exploiting the correlation between RS and total energy demand. 

\begin{table}[htb!]
    \centering
    \footnotesize
    \caption{Detection accuracy for reserves shortfall by functional depth metric. 
    }
    \setlength{\tabcolsep}{2.5pt} 
    \renewcommand{\arraystretch}{1.} 
    \begin{tabular}{r|rrrrrrrr}
        \toprule
        \textbf{Metric} & \textbf{ID} & \textbf{MBD} & \textbf{EXD} & \textbf{ERD} & \textbf{LID} & \textbf{HMD} & \textbf{DQ} & \textbf{RTD} \\
        \midrule
                \textbf{Min. [\%]} & 60 & 68 & 70 & 70 & 70 & \textbf{76} & 70 & 68 \\
        \textbf{Median [\%]} & 80 & 84 & 82 & 80 & 90 & \textbf{92} & 82 & 84 \\
        \textbf{Avg. [\%]} & 80.4 & 83.52 & 81.04 & 80.96 & 88 & \textbf{90.8} & 80.96 & 82 \\
                \textbf{Max. [\%]} & 94 & 96 & 94 & 94 & 98 & \textbf{100} & 92 & 94 \\
        \bottomrule
    \end{tabular}
    \label{tab:aggShort}
\end{table}

\begin{figure}[!htb]
    \centering
    \includegraphics[scale = 0.375, trim = {0cm, 0cm, 0cm, 0cm}, clip]{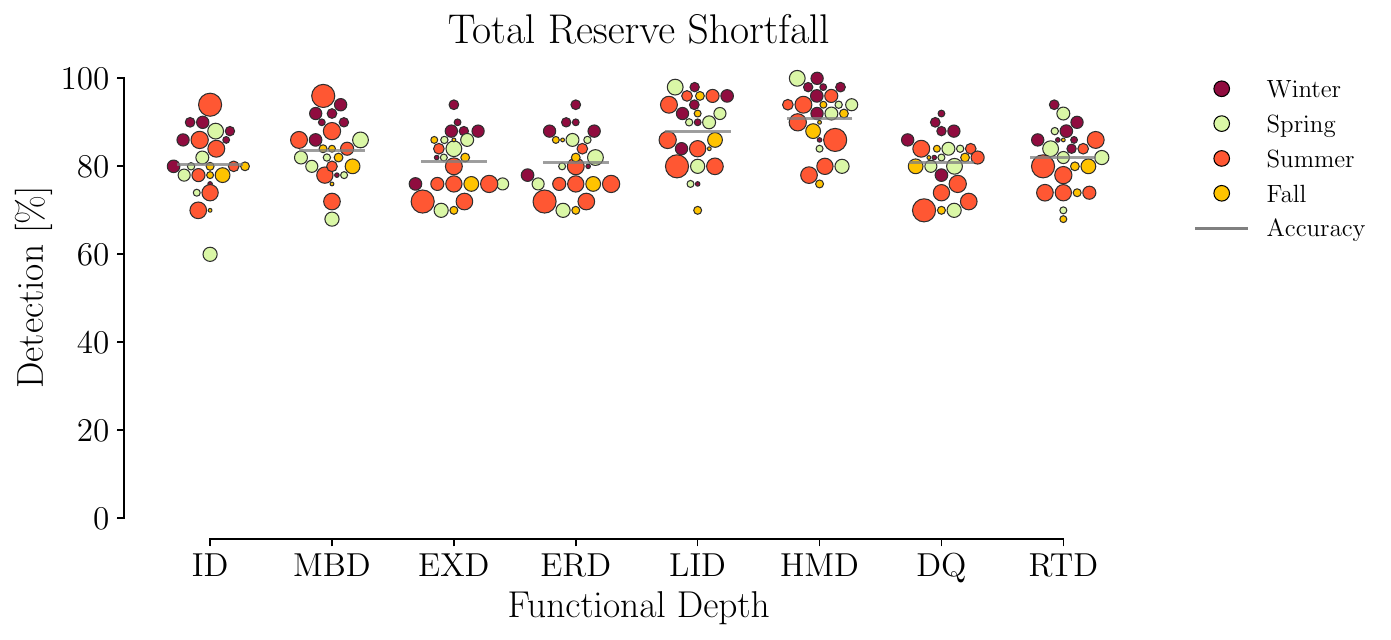}
    \caption{Detection accuracy for identifying RS. The bubble swarm plots correspond to different functional depth metrics ${\cal D}$(see Table~\ref{tab:aggShort}), with each dot representing the accuracy $P^{\cal D}_k$ achieved for the $25$ given simulation dates and the gray horizontal bars denoting the respective mean accuracy $Ave_k(P^{\cal D}_k)$. Symbol size denotes the magnitude of respective RS and the colors represent seasons of the year. }
    \label{fig:swarm-reserve}
\end{figure}

The summary in Table~\ref{tab:aggShort} shows that functional depth of aggregate NL efficiently detects scenarios with largest RS. The best functional depth method is HMD, followed by LID. 
In Fig.~\ref{fig:swarm-reserve}, we see that identifying extremality in RS is more challenging during summer when larger shortfalls (see size of red bubbles) tend to occur. Mean detection accuracy ranges from $80.4\%$ (ID) to 90.8\% (HMD). Given that the accuracy of AUC pre-screening without using any depth functions is 84.96\%,  only LID and HMD actually enhance overall detection accuracy. For many metrics, there is a skew in the distribution of $P_k$ reflecting the presence of a few poorly screened test days.


The choice of system NL as the (only) input for statistical depth is based on considering all alternatives (load, solar, wind, VRE generation, and net load) at the aggregated grid-level, and picking the one that works best. The pre-screening threshold $n_1$ in stage 1 is validated (considering it in increments of 50) to maximize detection accuracy; $n^{RS}_1=150$ is found to be best for predicting RS. 

\subsection{Load Shedding}\label{sec:load-shed}

We continue identifying scenarios that involve Load Shedding (LS). Unlike the previous section, most scenarios have zero LS $L_i = 0$, so the number of extreme scenarios with positive LS $L_i > 0$ is not fixed. Hence, our target set of operationally extreme scenarios ${\cal E}_k$ has a different cardinality for different test days $k=1,\ldots,K$. Specifically, on average there are $Ave_k(|{\cal E}_k|) = 149.48$ scenarios with LS. To preserve the same 50\% margin of misidentification, we therefore select on average $Ave_k(n_2(k)) \approx 225$ scenarios. 

Fig.~\ref{sm:scatter_rank_vs_shedding} in the supplementary material plots magnitude of LS against functional depth metrics. It shows that there is limited predictive pattern by looking at functional depth of grid-level NL, though we do observe a clear negative dependence, with the Kendall's $\tau$ rank correlation around $-0.35$ for LID and HMD. Yet there are many false positives (functionally extreme scenarios that lead to zero shedding) and many false positives (functionally deep scenarios that generate shedding). The resulting detection accuracy is about 75\% (LID yields 76.00\% followed by HMD at 75.28\%), compared to about 91\% accuracy for predicting RS in Table~\ref{tab:aggShort}. 

\begin{figure}[!htb]
    \centering
    \includegraphics[scale = 0.19, trim = {0cm, 0cm, 0cm, 0cm}, clip]{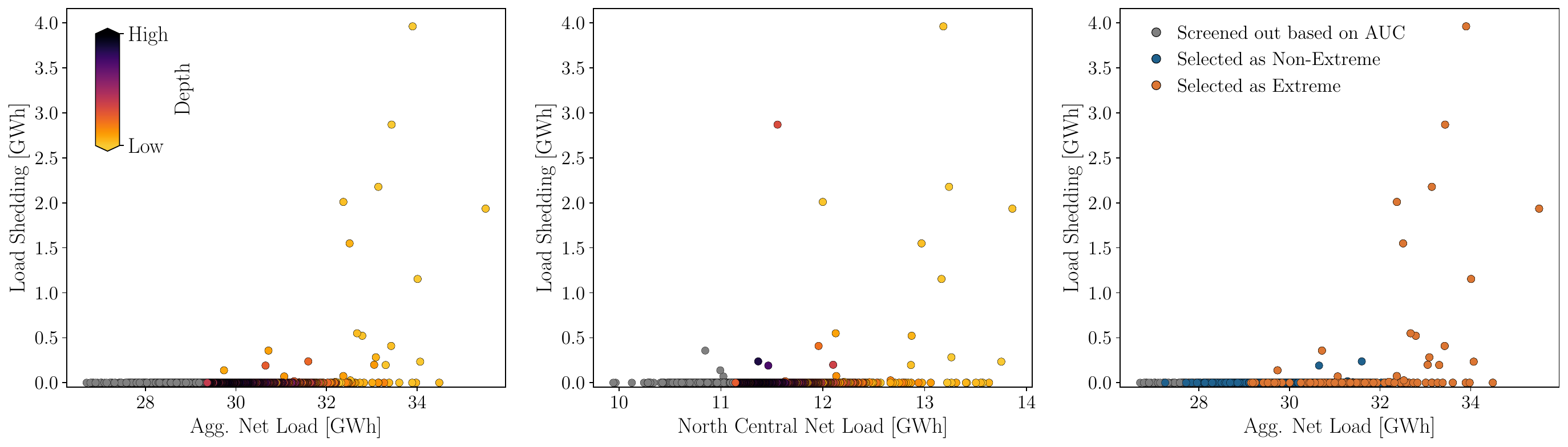}
    \caption{Three-stage detection of LS based on merging grid and zonal level functional depths on Feb~26, 2018. Hourly average NL on the $x$-axis and daily LS on the $y$-axis. \emph{Left}:  grid-level selection; \emph{Middle:} zonal-level selection. Scenarios screened out by NL AUC are in gray, the rest are color-coded (brighter is more extreme) according to ERD functional depth. \emph{Right:} merging of the two criteria: $n_2(k)=100$ most outlying scenarios in terms of combined depth ranking.}
    \label{fig:stepwise}
\end{figure}

The above lower detection accuracy is due to the fact that LS events are often caused by congestion. In the Texas-7k grid, loads are primarily in the Dallas-Fort Worth and Houston areas, while generation is more in West and North Texas. Thus LS is prone to arise during periods of high demand and high VRE generation that amplify this spatial mismatch. Such events cannot be identified using grid-level statistics. Motivated by above, we augment \emph{zonal}-level covariates to our predictors, using the ERCOT zone divisions. The pre-screening procedure followed in stage-1 does not require any adjustment in the methodology when including zonal-level covariates. The screening parameter $n_1$ is re-tuned to maximize detection accuracy, leading to $n^{LS}_1 = 550$. We then rank scenarios according to their functional depth both in terms grid-level NL and in terms of zonal NL. To merge the two depth scores, grid-wide $R^G_k$ and zonal $R^Z_k$, we select the scenario set ${\cal O}_k$ of size $n_2(k)$ based on the union  $\inf_j \{ | R^G_{(j),k} \bigcup R^Z_{(j),k} | \ge n_2 \}$, where $R_{(j)}$ refers to the $j$ scenarios with the least functional depth. Fig.~\ref{fig:stepwise} summarizes how this 3-stage selection works, showing the initial filtering by NL AUC, the computation of the depth metrics $R^G$ and $R^Z$, and finally their merging.

\begin{table}[htb!]
    \centering
    \scriptsize
    \caption{Accuracy of LS detection using selected functional depth metrics with adaptive number of selected scenarios $n_2(k)= |{\cal O}_k|$ across the 25 test days. $|{\cal E}_k|$ is the number of scenarios with realized LS and GWh refers to the total magnitude of load shedding (across scenarios) on that day. 
    }
    \setlength{\tabcolsep}{1.125pt} 
    \renewcommand{\arraystretch}{1.} 
    \begin{tabular}{c|rrr|rrrrrrrr}
        \toprule
        \textbf{Day} & $\mathbf{|{\cal E}_k|}$ & $\mathbf{|{\cal O}_k|}$ & \textbf{GWh} & \textbf{ID} & \textbf{MBD} & \textbf{EXD} & \textbf{ERD} & \textbf{LID} & \textbf{HMD} & \textbf{DQ} & \textbf{RTD} \\
        \midrule
        01-02 & 681 & 678 & 2574 & 2560 & 2561 & 2557 & 2557 & 2559 & 2560 & 2551 & 2556 \\
        01-20 & 41 & 100 & 31.50 & 27.51 & 29.12 & 30.73 & 30.73 & 29.91 & 29.91 & 31.08 & 27.10 \\
        02-13 & 55 & 100 & 35.92 & 16.58 & 17.36 & 34.24 & 34.24 & 33.91 & 32.69 & 34.36 & 33.64 \\
        02-14 & 50 & 100 & 78.58 & 74.24 & 75.16 & 77.41 & 77.89 & 77.59 & 78.04 & 77.99 & 76.57 \\
        02-26 & 24 & 100 & 19.17 & 18.51 & 18.52 & 18.73 & 18.73 & 18.65 & 18.59 & 18.73 & 18.73 \\
        03-14 & 75 & 100 & 72.87 & 45.16 & 46.17 & 63.92 & 63.92 & 59.14 & 55.83 & 65.73 & 56.29 \\
        04-01 & 131 & 100 & 108.30 & 41.35 & 39.65 & 51.73 & 51.73 & 37.98 & 37.19 & 44.93 & 41.18 \\
        04-09 & 61 & 100 & 39.31 & 13.57 & 14.25 & 28.16 & 28.16 & 31.56 & 26.92 & 28.53 & 15.73 \\
        04-24 & 159 & 100 & 127.52 & 58.80 & 63.97 & 78.45 & 76.51 & 81.18 & 73.36 & 79.70 & 66.90 \\
        05-10 & 99 & 100 & 54.40 & 23.31 & 22.54 & 28.98 & 28.80 & 20.40 & 23.52 & 27.73 & 18.35 \\
        05-25 & 166 & 100 & 157.49 & 77.29 & 77.51 & 81.67 & 82.35 & 68.21 & 71.63 & 79.34 & 56.67 \\
        06-04 & 72 & 102 & 40.98 & 21.04 & 21.06 & 31.79 & 31.79 & 19.52 & 21.39 & 31.75 & 19.73 \\
        06-30 & 89 & 102 & 46.49 & 20.81 & 23.34 & 31.55 & 29.40 & 23.93 & 27.72 & 29.92 & 26.02 \\
        07-22 & 855 & 1000 & 2106 & 2106 & 2106 & 2106 & 2106 & 2106 & 2106 & 2106 & 2106 \\
        07-24 & 198 & 953 & 109.14 & 109.14 & 109.14 & 109.14 & 109.14 & 109.14 & 109.14 & 109.06 & 109.14 \\
        08-08 & 213 & 752 & 99.77 & 97.87 & 97.87 & 98.13 & 98.13 & 97.72 & 98.36 & 98.13 & 97.41 \\
        08-18 & 42 & 252 & 12.20 & 10.46 & 10.66 & 11.92 & 11.92 & 10.57 & 10.58 & 11.92 & 10.16 \\
        09-04 & 7 & 100 & 2.13 & 0.55 & 0.55 & 0.84 & 0.84 & 0.84 & 0.55 & 0.84 & 0.33 \\
        09-14 & 49 & 100 & 22.29 & 12.48 & 12.48 & 14.46 & 13.11 & 13.96 & 11.00 & 13.92 & 9.11 \\
        10-02 & 118 & 100 & 64.83 & 36.51 & 36.48 & 40.61 & 40.72 & 38.57 & 44.19 & 37.13 & 32.57 \\
        10-17 & 96 & 100 & 1.63 & 0.19 & 0.19 & 0.49 & 0.49 & 0.48 & 0.17 & 0.45 & 0.16 \\
        11-02 & 34 & 100 & 5.00 & 2.00 & 3.87 & 4.00 & 4.00 & 4.00 & 4.02 & 4.00 & 4.02 \\
        11-13 & 179 & 100 & 60.48 & 39.16 & 40.44 & 43.25 & 43.42 & 46.94 & 47.83 & 44.57 & 35.48 \\
        12-01 & 69 & 100 & 38.90 & 30.23 & 30.20 & 35.49 & 35.49 & 30.46 & 30.98 & 35.97 & 29.99 \\
        12-27 & 174 & 100 & 217.59 & 155.04 & 157.88 & 164.04 & 163.43 & 174.09 & 165.44 & 166.98 & 160.93 \\
        \midrule
        \textbf{Total} & 3737 & 5639 & 6126 & 5598 & 5615 & 5742 & 5737 & 5693 & 5685 & 5730 & 5607 \\
        \textbf{Avg.} & 149.5 & 225.6 & 245.04 & 223.92 & 224.60 & 229.68 & 229.49 & 227.72 & 227.40 & 229.19 & 224.30 \\
        \midrule
        \multicolumn{4}{c}{\textbf{Accuracy [\%]} } & 91.38 & 91.66 & \textbf{93.73} & 93.65 & 92.93 & 92.80 & 93.53 & 91.53 \\
\bottomrule
    \end{tabular}
    \label{tab:aggZonalShed}
\end{table}

\begin{figure}[!htb]
    \centering
    \includegraphics[scale = 0.255, trim = {0cm, 0cm, 0cm, 0.2cm}, clip]{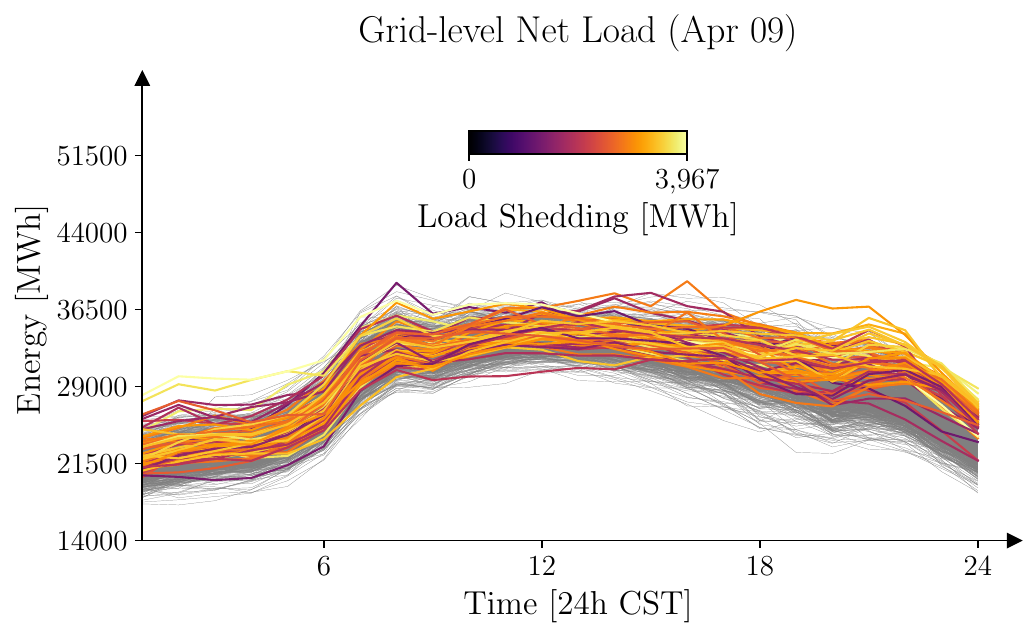}
    \includegraphics[scale = 0.255, trim = {0.9cm, 0cm, 0cm, 0.2cm}, clip]{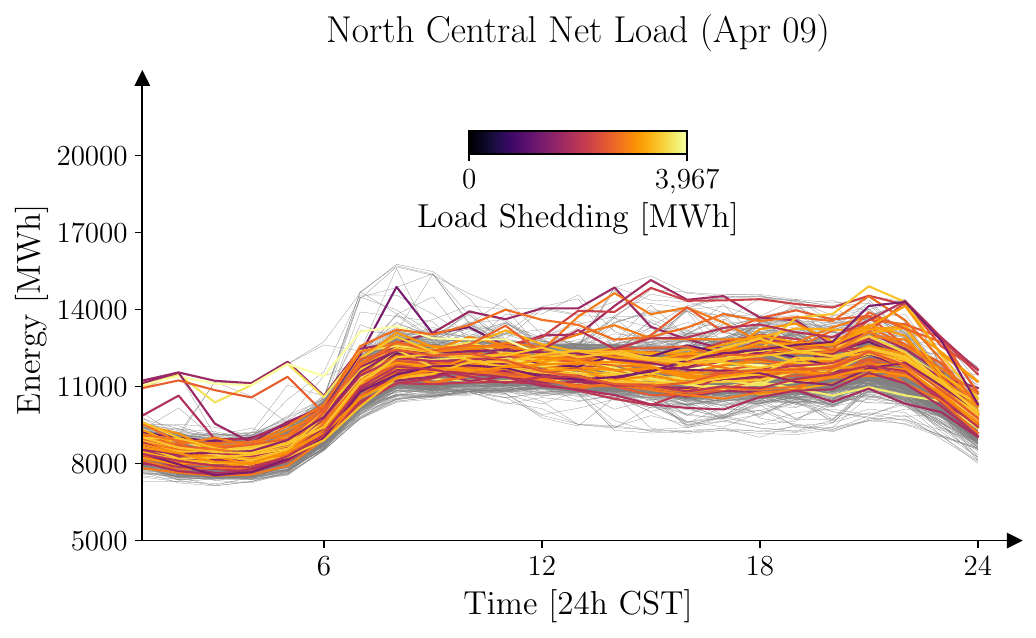}
    \caption{Selecting scenarios with load shedding on Apr~9, 2018. Scenarios with zero LS are in gray, otherwise brighter color indicates more LS. The proposed selection method applies functional depth to the grid-level NL (in left panel) and the NC-zonal NL (right).}
    \label{fig:shed_extreme}
\end{figure}

To come up with this recipe, we evaluated the performance of each zonal facet in combination with grid-level NL. The possible zonal facets (load, solar/wind/VRE generation, and NL) were added as a predictor one at a time. Among these 40 predictors (8 ERCOT zones times 5 facet types), North Central (NC) NL was found to be the best, see Fig.~\ref{fig:shed_extreme} and Fig.~\ref{fig:stepwise}). Fig.~\ref{fig:shed_extreme} shows that extreme LS scenarios  are linked to high grid-level NL during the morning peak and high NC-zonal NL especially in the evening. 

In the second stage, we select an adaptive number of scenarios $n_2(k)$, determined by the overall level of demand for day $k$. Since LS primarily occurs on days with high NL, we link $n_2(k)$ to the number of scenarios with grid-level peak NL exceeding $62.5$~GWh, $n_2(k) := 100 + | i : \max_t \hat{f}^{\cal N}_{i,k}(t) \geq 62.5 |$. 
On a couple of test days with very high peak NL, we have $n_2(k) > n_1$, in which case we just pre-screen $n_2(k)$ scenarios based on AUC NL. 


The best performance for screening for LS is 93.73\% achieved by EXD using adaptive $n_2(k)$, $n^{LSZ}_1=650$ and NC-zone NL. DQ and ERD perform almost as well. Adaptively sizing $n_2(k)=|{ \cal O}_k|$ is a key ``trick'', as accuracy drops to only 77.47\% if one takes $n_2$ to be constant, see Tables~\ref{sm:aggShed}~and~\ref{sm:aggZonalShed_noAdap}. 


\begin{figure}[!htb]
    \centering
    \includegraphics[scale = 0.375, trim = {0cm, 0cm, 0cm, 0cm}, clip]{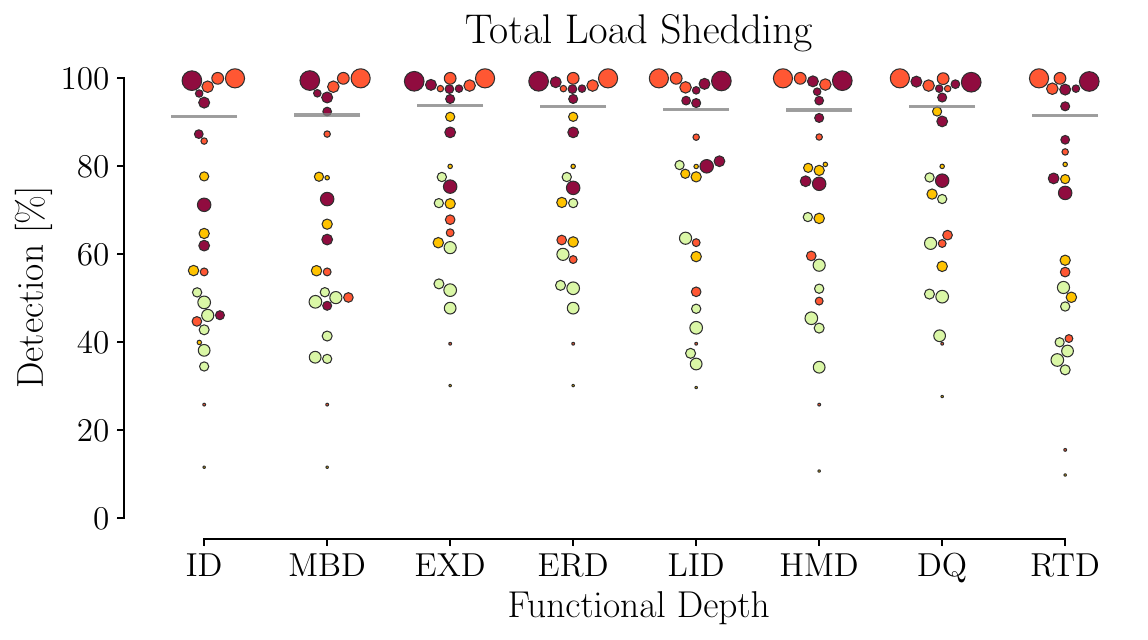}
    \caption{Detection accuracy for identifying LS 
    using zonal information and adaptive $n_2(k)$. The bubble swarm plots correspond to different functional depth metrics (see Table~\ref{tab:aggShort}), with each dot representing the accuracy $P^{\cal D}_k$ achieved across the 25 test days (gray bar shows average accuracy). Bubble size encodes the magnitude of LS on that day; colors represent winter (purple), spring (green), summer (red), and fall (yellow).}
    \label{fig:shed_fdepth}
\end{figure}

\subsection{VRE Curtailment}\label{sec:curtailment}

For our last selection task, we aim to identify scenarios that lead to VRE curtailment (VC). This is the hardest task, since VC is highly sensitive to the grid topology. Consequently, manual detection of high-curtailment scenarios is infeasible and an automated algorithmic procedure like ours is essential. In the considered Texas-7k grid, during times of high electricity demand \emph{and} high VRE generation, the transmission lines and transformers are susceptible to congestion which in turn can lead to localized curtailment. Thus curtailment events may be triggered by asset-specific, zonal or grid-wide configurations. 

On some days, there is no VC across all scenarios, which makes the prediction moot; we exclude these days in our analysis. Conversely, on some days nearly all scenarios have VC, due to the strong spatial concentration of VRE generators which leads to economic congestion in that zone. We exclude those days (namely days with more than 250 VC scenarios) as well, since there is no meaningful down-selection to do.

Since curtailed energy is often very low, we define operational extremality as scenarios with daily curtailment of $\geq 100$ MWh. With this restriction, there are on average 50.39 extreme VC scenarios across the 18 test days that we are left with. Preserving a 50\% margin, we select $n_2 = \lceil 1.5 \cdot 50.39 \rceil \approx 76$ scenarios each day.
%
%
%
Because VC events do not correlate with either high energy demand or high  solar and/or wind generation (see suppl.~Fig.~\ref{sm:curt_extreme_load}) we find that screening them by AUC is no longer beneficial, and switch back to a single-stage selection procedure. 

\begin{figure}[!htb]
    \centering
    \includegraphics[scale = 0.375, trim = {0cm, 0cm, 6.5cm, 0cm}, clip]{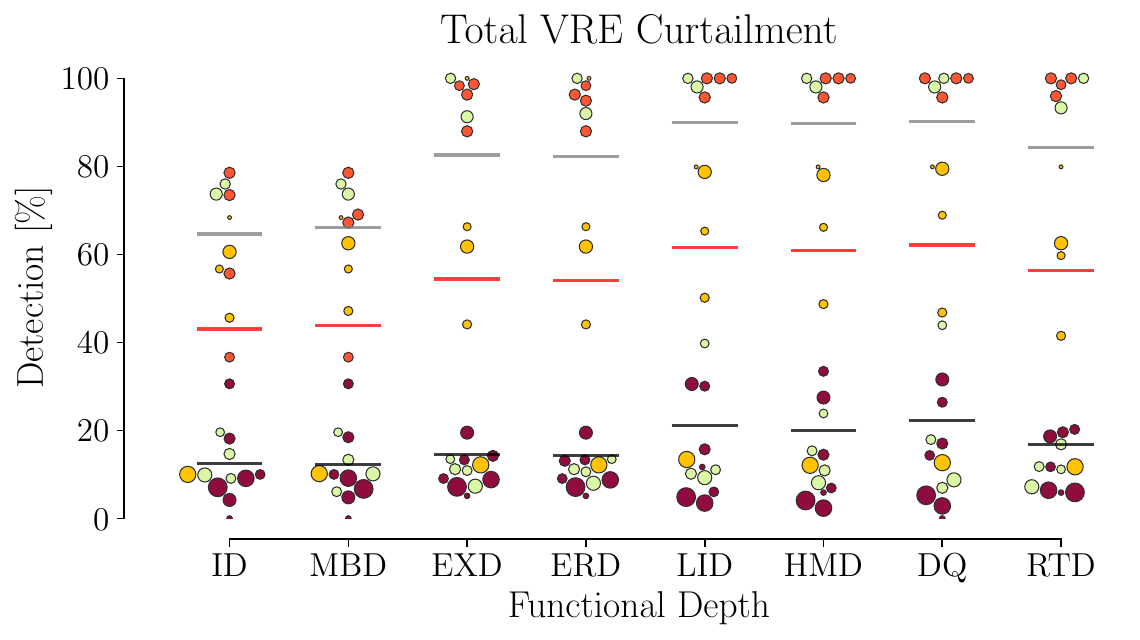}
    \caption{VRE curtailment detection accuracy across different functional depth metrics, 
    see Table~\ref{sm:aggZonalCurtail_short}. 
    Bubble size encodes magnitude of VC on the respective day, and color encodes season of the year (winter: purple; spring: green; summer: red; fall: yellow). Red lines indicate average accuracy $Ave_k( P^{\cal D}_k)$ across the 18 test days. }
    \label{fig:curt_fdepth}
\end{figure}

Given the context, it is not surprising that VC prediction is challenging. Using grid-level VRE generation and NL as predictors, average accuracy of less than 50\% is achieved, ranging from 44.69\% for DQ (see Fig.~\ref{fig:curt_fdepth}, where EXD and ERD show very similar performance) to 37.82\% for ID. A slight improvement is achieved by augmenting with zonal information: the best functional depth metrics are LID and HMD, with similar accuracy of 61.58\% and 60.91\% respectively (see Table~\ref{sm:aggZonalCurtail_short}).

\begin{figure}[!htb]
    \centering
    \includegraphics[scale = 0.255, trim = {0cm, 0cm, 0cm, 0cm}, clip]{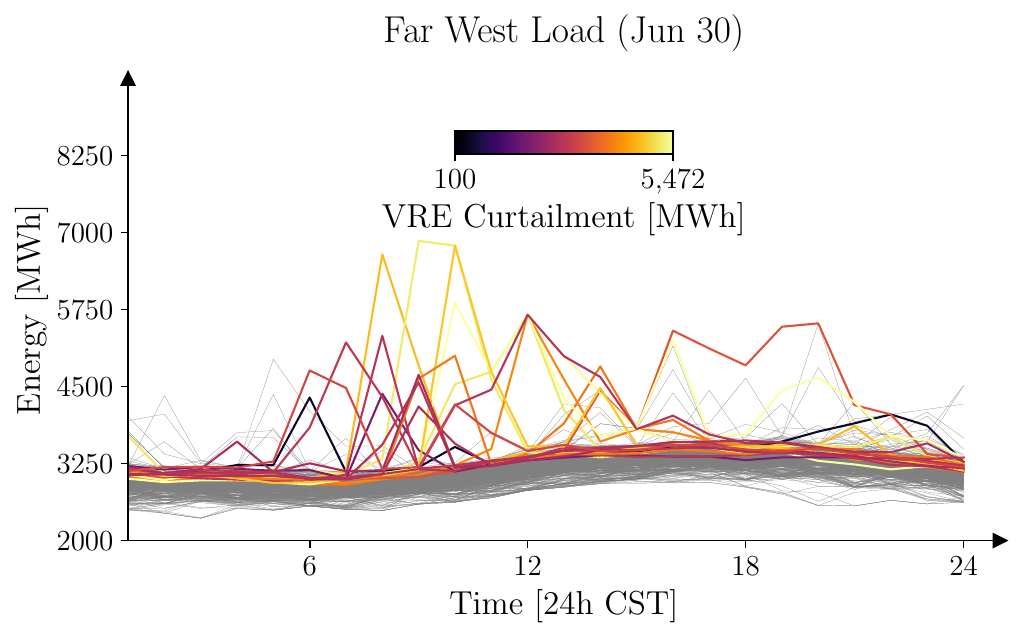}
    \includegraphics[scale = 0.255, trim = {0.9cm, 0cm, 0cm, 0cm}, clip]{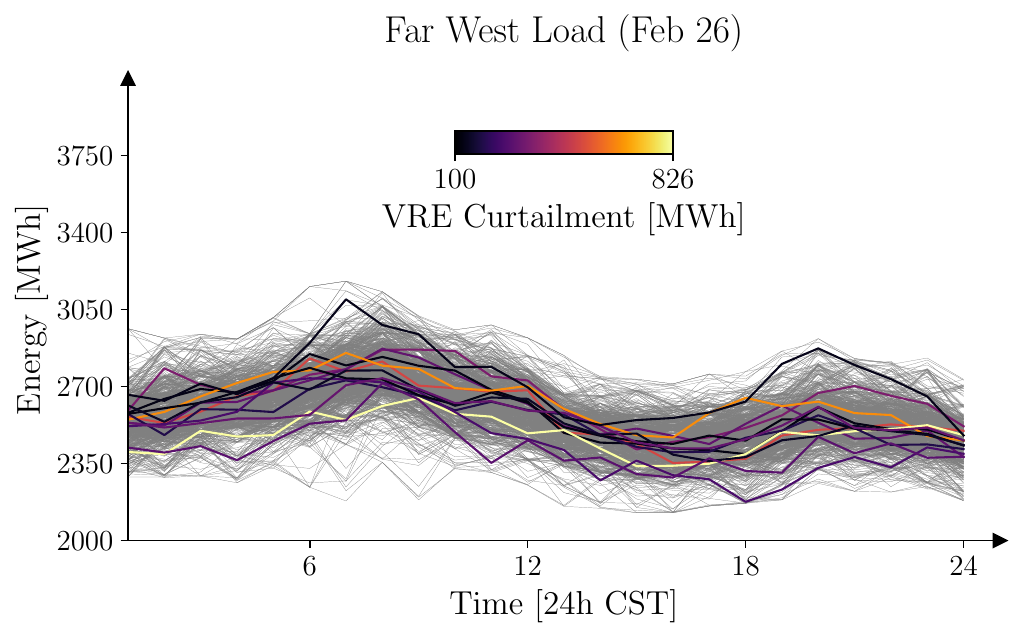}
    \caption{Scenarios with VRE curtailment on Jun~30 and Feb~26, 2018. The brighter the color gradient, the larger the magnitude of VC in that scenario. The left panel shows that Far West zone load outlyingness is predictive of VC in the hotter months. However, the right panel shows that this is no longer the case in winter. }
    \label{fig:curt_extreme}
\end{figure}

The results in Fig.~\ref{fig:curt_fdepth} show a bimodal distribution that is closely correlated to seasons of the year, clustering according to a warm weather (summer-fall)/cold weather (winter-spring) pattern. This seasonality is related to the shifting daily load profile, namely a mid-day peak that occurs during the hotter months due to air conditioning demand, see supplementary Fig.~\ref{sm:curt_extreme_load}. To exploit this pattern, we consider seasonal zonal-level predictors which allow further gains. In particular, we find that FW-zone load is a predictor of VC during summer/fall. Fig.~\ref{fig:curt_extreme} suggests that one underlying driver is strong down-ramps in load that cause curtailment of wind generation as the least-cost regulation mechanism. Best performance in warm months is obtained by pre-screening to keep $n^{VCZ}_1=450$ scenarios according to grid-level NL AUC, and then applying either DQ (accuracy of 90.19\%) or LID (90.06\%). None of the proposed depth metrics help for cold-weather VC, which appears to require drilling down to asset-level scenario features. 

\section{Discussion}\label{sec:discuss}

\subsection{Directionality}

Operational extremality is \emph{directional}, e.g.~we are interested in scenarios with exceptionally high RS, but not with exceptionally low shortfall. The presented depth functions are however two-sided, i.e.~focus on outliers relative to a direction-agnostic center-outward ranking. This creates a mis-match to the extent that the scenarios also have a top-down structure, e.g.~scenarios with high load are more likely to be extreme, (see Figs.~\ref{fig:agg_net_load_extreme}~and~\ref{fig:shed_extreme}). Above we used the AUC to pre-screen scenarios to resolve this issue. 
Pre-screening is optimized with a task-dependent $n_1$. For RS, we keep only $n_1^{RS}=150$ scenarios, due to the strong correlation with net load AUC. For LS we have $n_1^{LS}=550$ and $n_1^{LSZ} = 650$ when including NC-zonal NL. In contrast, VC only benefits from pre-screening during the summer/fall months ($n_1^{VCZ} = 450$), when curtailment is highly correlated to FW-zonal high load.


Another alternative to AUC screening is to to redefine the depth notion to construct one-sided orderings. Taking load $\hat{f}^{\cal L}$ as an example, we would like to only select scenarios with ``large value extremality", i.e.~pointwise ranks close to 1. Such one-sided versions are possible for MBD, DQ and ERLD. For instance, in ERLD one could replace $d_i(t)$ with $R_i(t)$ in \eqref{eq:ERLD}, sorting scenarios by their univariate ranks (top is more extreme), rather than by univariate depth. Similarly, one can modify the stochastic order underlying RTD or EXD to create a one-sided implementation. On the other hand, one-sided versions do not make sense for LID and HMD, where the distance-based construction destroys directionality. 

\subsection{Grid Properties}

Our analysis demonstrates that extreme events do not only depend on the energy demand and availability of VRE resources but also on  grid topology. 
Congestion in transmission lines and transformers is the principal cause of VC during the warm weather season in Texas-7k (see Fig.~\ref{sm:vatic_simulations}). In contrast, in cold months VC is more linked to the more typical combination of low load and high VRE generation. These different patterns cause the multimodal distribution in Fig..~\ref{fig:curt_fdepth} and highlight the complex interaction between high-level net load profiles and low-level, seasonally-driven grid properties.



Similarly, the most challenging season to foresee RS is summer, which indicates that the uncertainty in the energy demand and VRE generation is highest during this season suggesting the value of non-constant operational reserve margins.

\subsection{Which Depth Function?}


The presented depth functions have different merits across the facets of operational extremality. The four best-performing are LID and HMD, which are distance-based, and EXD and ERD which are rank-based. LID does best at detecting RS, followed by HMD. EXD and ERD achieve highest accuracy for detecting LS. For VC, DQ performs best, followed by LID. We recommend EXD and ERD which appear to be more stable than the distance-based functional depths. 

In contrast, ID, MBD and RTD perform relatively poorly. Their accuracy across all case studies is not as good, and they tend to be less stable, manifesting as bigger spread and/or multi-modality of $P_k$ across test days.



\section{Conclusion}\label{sec:conclude}

The above results demonstrate that statistical measures of functional depth offer a novel and effective approach to screen scenarios for operational risk assessment. Rather than having to simulate ED operation across \emph{all} scenarios, the proposed methodology offers a ranking of scenarios that closely matches their extremality in terms of variable generation costs, reserve shortfall, Load shedding, and Variable Renewable Energy curtailment. Thus, the effort needed to identify the most risky scenarios can be cut down by an order of magnitude, and in turn help to implement better planning during the day-ahead unit commitment phase. 

Our results also contain evidence for the localized drivers of LS and VRE curtailment. These events tend to be caused by congestion, and hence cannot be identified by looking at aggregate load and VRE generation. Additional analysis and approaches are therefore warranted for these tasks; since the considered scenarios are at asset-level, more fine-grained screening strategies could in principle be implemented.

Performance gains are likely possible through further customization of the functional depth measures. One starting point would be to implement one-sided variants of some of the depth metrics. A complementary extension would be to consider functional depths of multiple scenario facets and then combine them in a more sophisticated way. It remains an open problem how to effectively merge multiple depth rankings to come up with the final subset of selected scenarios. We have proposed a simple ``union'' of two different rankings in Section~\ref{sec:load-shed}; more general multivariate variants are left for future research. Another alternative route would be to explore statistical learning approaches to exploit the multi-modality in the data (i.e.~mixture models) and/or annual seasonality.

\section*{Data and Software Availability}

\textit{Vatic} is publicly available for Python programming language at  \url{https://github.com/PrincetonUniversity/Vatic}. The analysis in this investigation was performed in Python and R; code is available at \url{https://github.com/gterren/ranking_extreme_scenarios}. Depth metrics are available in the R libraries \textit{fdaoutlier} \cite{fdaoutlier} and \textit{fda.usc} \cite{fda.usc}. \textit{Texas-7k} grid specification is at \url{https://electricgrids.engr.tamu.edu/texas7k}.


\section*{Acknowledgments}

This research was partially founded by the ARPA-E PERFORM grant DE-AR0001289. Use was made of computational facilities of the UCSB Center for Scientific Computing funded by 
CNS-1725797 and supported by MRSEC (NSF DMR 2308708). 
We thank X.~Yang, G.~Swindle, R.~Sircar and R.~Carmona for insightful discussions, and M.~Grzadkowski, A.~Fang for computing support in deploying \emph{Vatic}.

\bibliographystyle{IEEEtran}

\bibliography{biblio_extreme}

\section{Biography Section}
 

\vspace{-33pt}

\begin{IEEEbiography}
[{\includegraphics[width=1in,height=1.25in,clip,keepaspectratio]{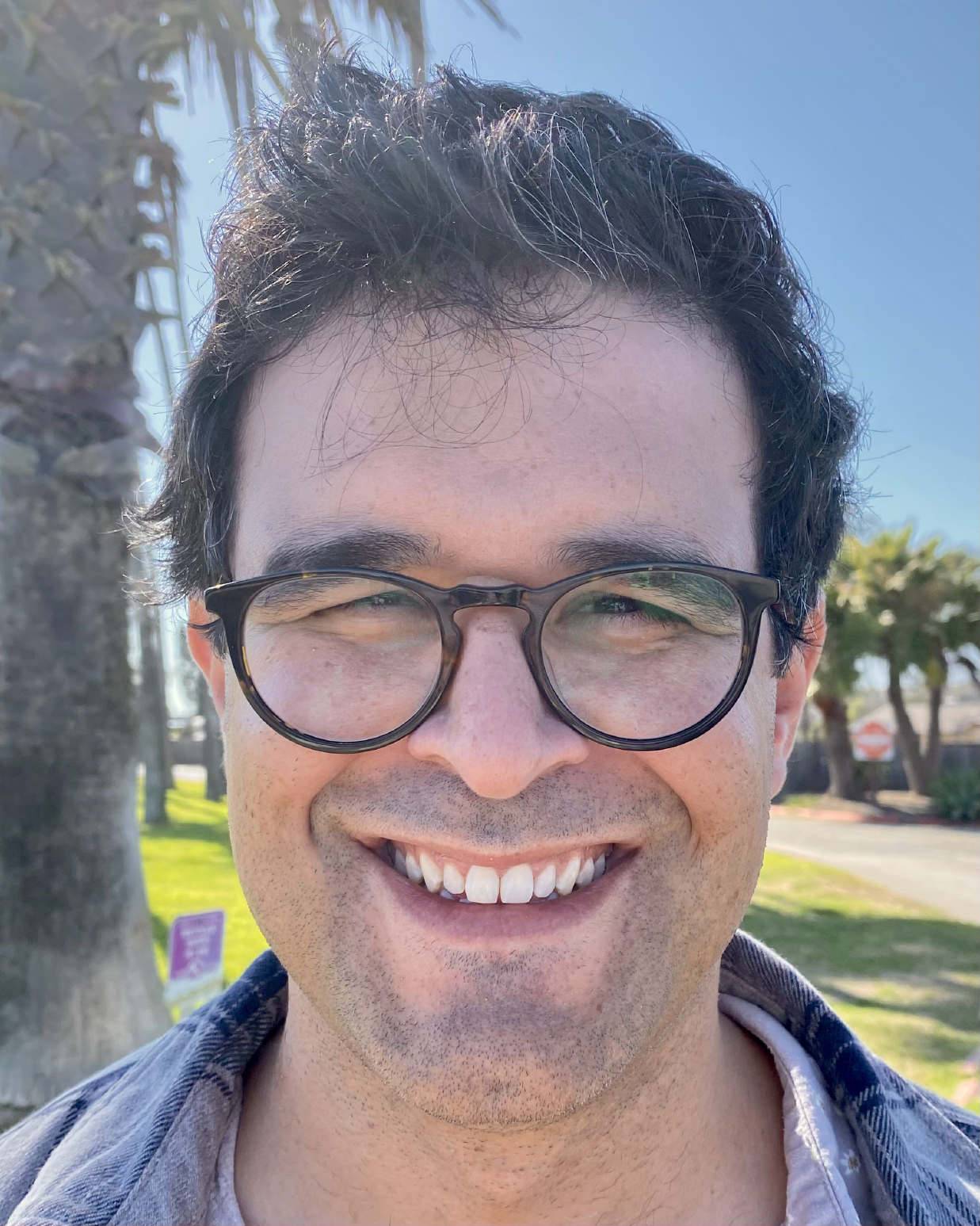}}]{Guillermo Terrén-Serrano} is a Postdoctoral Climate Innovation Fellow at the University of California Santa Barbara in Santa Barbara, CA, USA. He received a Bachelor's degree in Technical Industrial Engineering from the Universidad de Zaragoza in Zaragoza, Spain, and a Ph.D.~in Engineering from the department of Electrical and Computer Engineering at the University of New Mexico in Albuquerque, NM, USA. His research focus on the estimation of operating reserves and the planning reserve margin in macro-electricity systems, smart grids, and micro grids, using probabilistic machine learning, computer vision, and remote sensing.
\end{IEEEbiography}

\begin{IEEEbiography}               
[{\includegraphics[width=1in,height=1.25in,keepaspectratio,clip,trim=0in 0.45in 0in 0in]{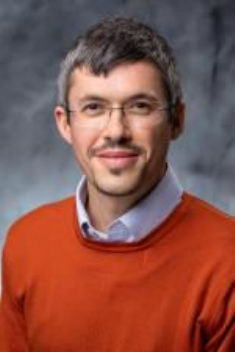}}]{Mike Ludkovski} is a Professor of Statistics \& Applied Probability at the University of California Santa Barbara in Santa Barbara, CA, USA. He received a Bachelor's degree in Mathematics from Simon Fraser University in Burnaby, Canada and a Ph.D.~in Operations Research and Financial Engineering from Princeton University in Princeton, NJ, USA. His research interest include stochastic control and stochastic modeling. Recently he has been working on statistical analysis of renewable generation and its short-term uncertainty quantification.
\end{IEEEbiography}

{\appendices
\begin{figure*}[!htb]
    \centering
    \includegraphics[scale = 1.125, trim = {0cm, 0cm, 0cm, 0cm}, clip]{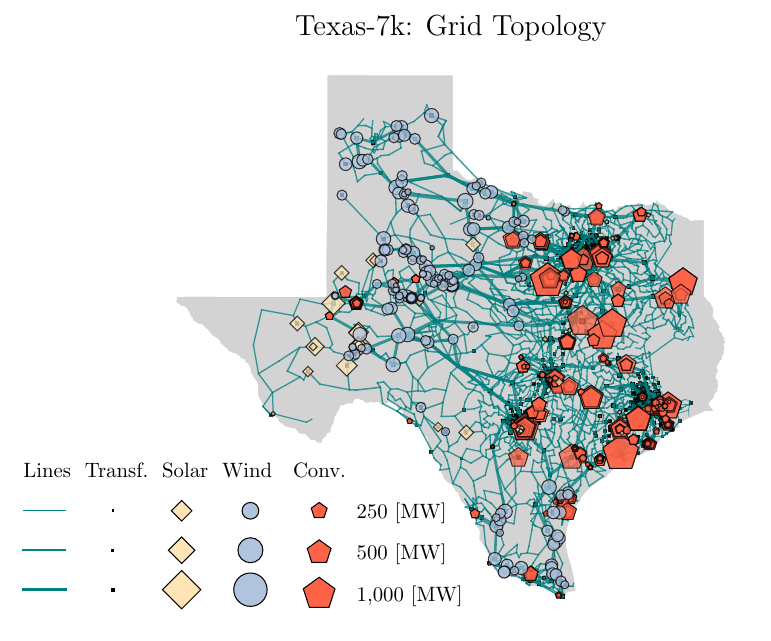}
    \caption{Topology of Texas-7k power grid. The devices in the power grid include 7,173 lines (dark and light green), 1,967 transformers (black), and 36 solar (yellow), 153 wind (blue) and 542 convectional generators (red).}
    \label{sm:topology}
\end{figure*}

\begin{figure*}[htb!]
    \centering
    \includegraphics[scale = 0.735, trim = {0cm, 0cm, 0cm, 0cm}, clip]{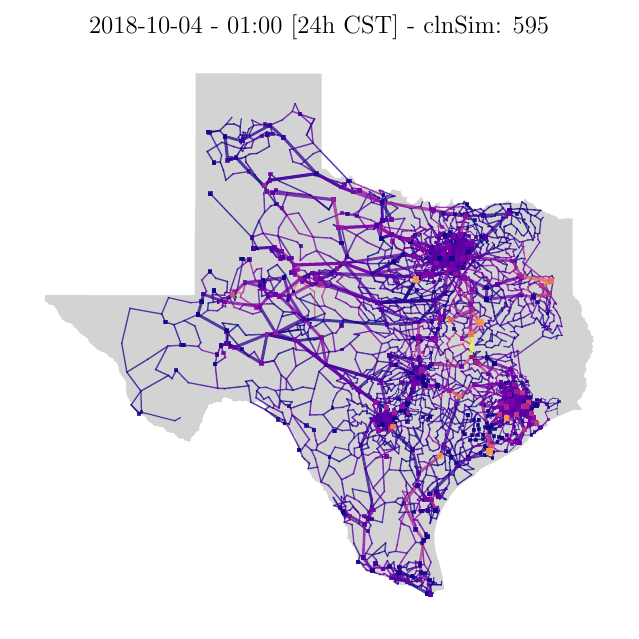}
    \includegraphics[scale = 0.735, trim = {0cm, 0cm, 0cm, 0cm}, clip]{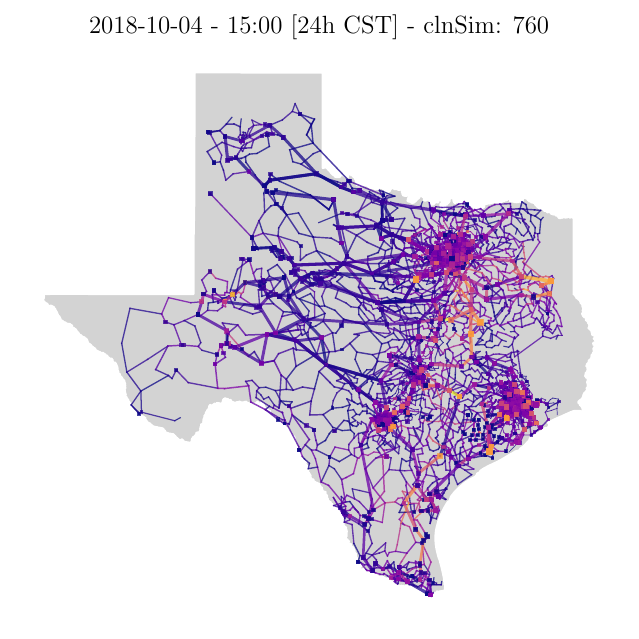}
    \includegraphics[scale = 0.735, trim = {0cm, 0cm, 0cm, 1cm}, clip]{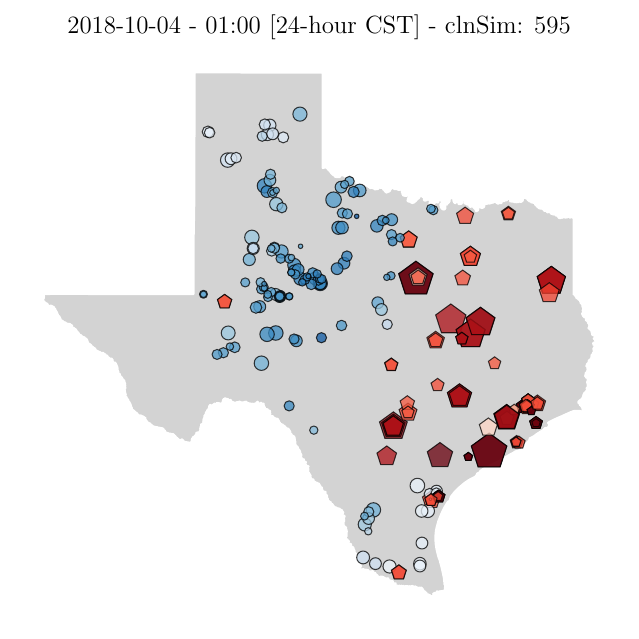}
    \includegraphics[scale = 0.735, trim = {0cm, 0cm, 0cm, 1cm}, clip]{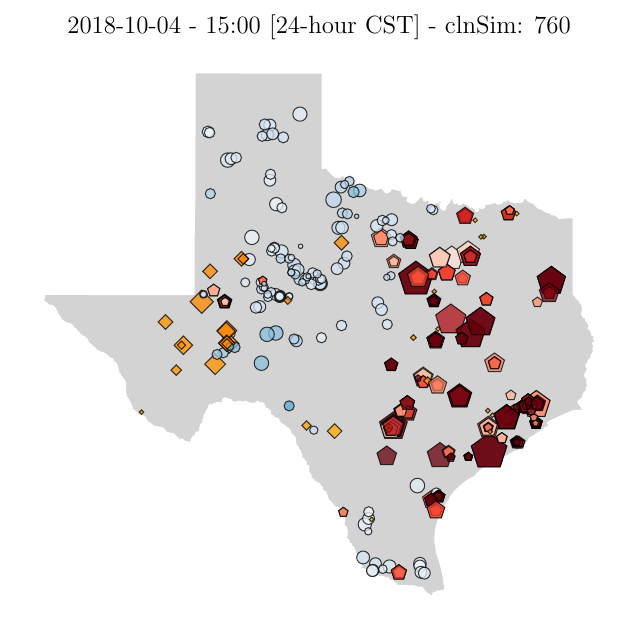}
    \includegraphics[scale = 0.635, trim = {1.15cm, 7cm, 3cm, 0.15cm}, clip]{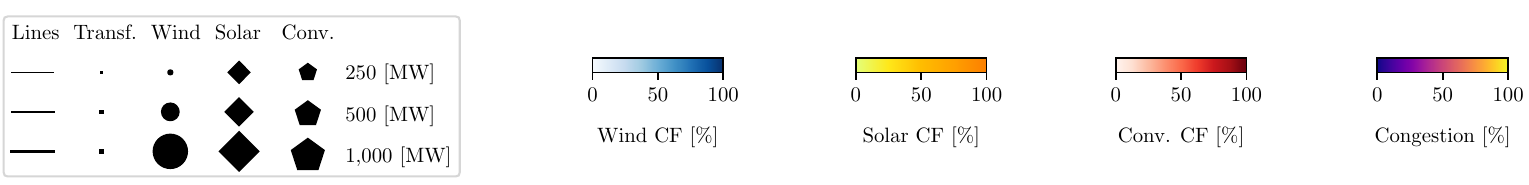}
    \caption{Operation simulation on Oct~4, 2018. This day has VRE energy curtailment at 1 am in scenario 595, and load shedding at 3 pm in scenario 760. In the VRE curtailment event, a line in east Texas (see bottom left) creates a load pocket that prevents the flow of energy from high wind generation in north west Texas (see top left map). In contrast, the load shedding event is triggered by a low wind generation which produces to overload a line (see top right map) preventing the energy generated from conventional assets to flow to north central region (see bottom right map).}
    \label{sm:vatic_simulations}
\end{figure*}

\begin{figure*}[!htb]
    \centering
   \includegraphics[scale = .45, trim = {0cm, 0cm, 0cm, 0cm}, clip]{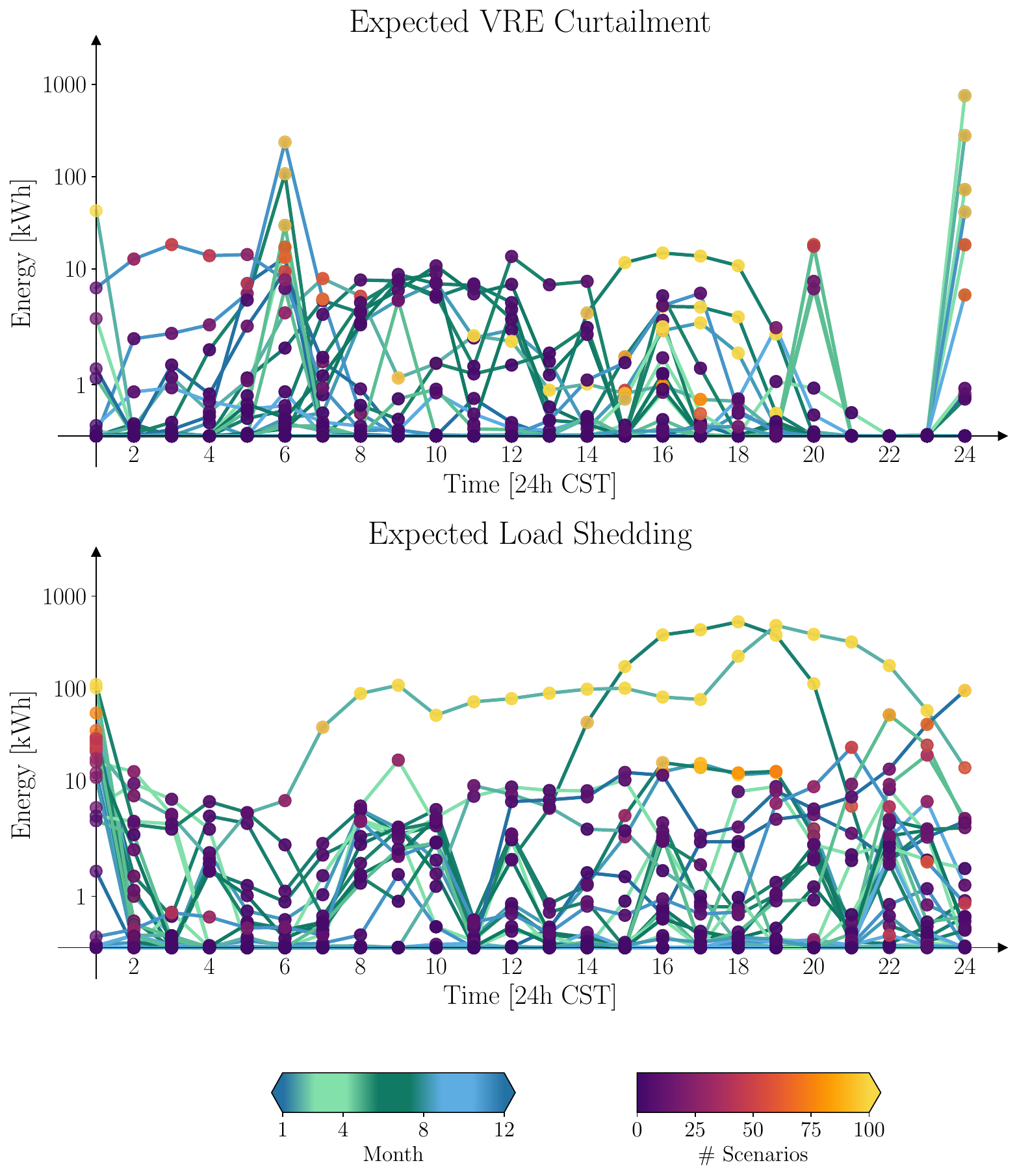}
    \caption{Magnitude of load shedding (top) and VRE curtailment (bottom) across scenarios. Each curve corresponds to one of the 25 simulated days. The  color gradient of the markers encodes the frequency of $L_i(t) >0$ or $C_i(t) >0$ (the brighter the more scenarios). Line color indicates the season of that simulated day (winter light colors and summer dark).}
    \label{sm:load_shedding_and_energy_curtailment}
\end{figure*}

\begin{figure*}[!htb]
    \centering
    \includegraphics[scale = 0.315, trim = {0cm, 0cm, 0cm, 0cm}, clip]{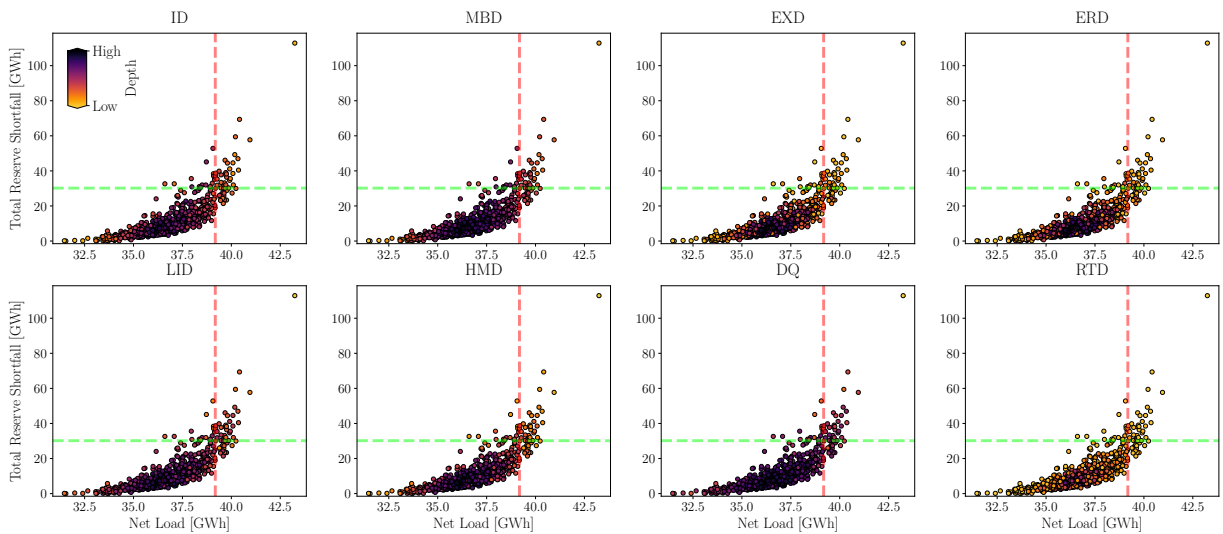}
    \caption{Detecting reserves shortfall via different functional depth metrics on Feb~13, 2018. Aggregated daily NL on the $x$-axis vs.~daily RS on the $y$-axis. Scenarios are color coded according to the respective depth metric: the brighter the color gradient, the less deep the scenario. Green horizontal (vertical red) line at 30.17~GWh (resp.~39.18 GWh) shows the threshold for the top 50 highest RS (resp.~top 50 highest NL).}
    \label{sm:scatter_net_vs_shortfall}
\end{figure*}

\begin{figure*}[!htb]
    \centering
    \includegraphics[scale = 0.315, trim = {0cm, 0cm, 0cm, 0cm}, clip]{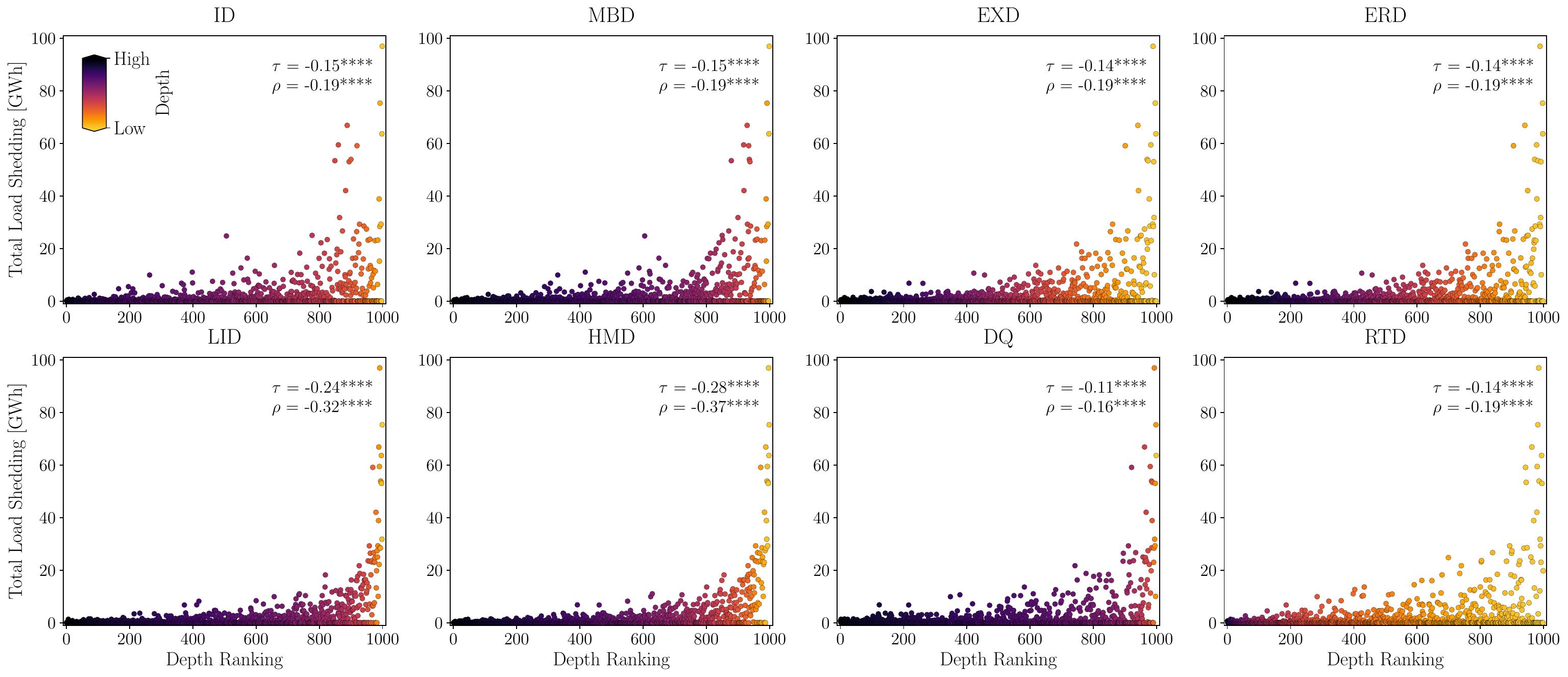}
    \caption{Load shedding aggregated at grid-level across the 24 hours on the $y$-axis vis-a-vis the depth ranking on the $x$-axis for different functional depth metrics on Jun~2, 2018. The more extreme a scenario, the higher its rank, thus the lower its depth score. Brighter color means a higher depth score. Most extreme scenarios appear on the right-hand side of the graphs. $\tau$ denotes the  Kendall's correlation coefficient, and $\rho$ is the Spearman's correlation coefficient. Their $p$-values are summarized with asterisks.}
    \label{sm:scatter_rank_vs_shedding}
\end{figure*}

\begin{table*}[htb!]
    \centering
    \small
    \caption{Accuracy of LS detection using selected functional depth metrics. We select $n_2= |{\cal O}_k|= 225$ scenarios across the 25 test days. $|{\cal E}_k|$ is the number of scenarios with realized LS and GWh refers to the total magnitude of load shedding (across scenarios) on that day. We first screen We select 225 scenarios by first screening 550 scenarios based on AUC NL and then using functional depth on grid-wide NL. } 
    \setlength{\tabcolsep}{2.5pt} 
    \renewcommand{\arraystretch}{1.} 
    \begin{tabular}{c|rrr|rrrrrrrr}
    \toprule
        \textbf{Day} & $|{\cal E}_k|$ & $|{\cal O}_k|$ & \textbf{GWh} & \textbf{ID} & \textbf{MBD} & \textbf{EXD} & \textbf{ERD} & \textbf{LID} & \textbf{HMD} & \textbf{DQ} & \textbf{RTD} \\
        \midrule
        01-02 & 681 & 225 & 2573.91 & 2141.05 & 2188.29 & 2183.83 & 2182.45 & 2224.03 & 2223.62 & 2163.23 & 2041.24 \\
        01-20 & 41 & 225 & 31.50 & 31.14 & 31.14 & 31.40 & 31.40 & 31.39 & 31.44 & 31.40 & 31.44 \\
        02-13 & 55 & 225 & 35.92 & 33.04 & 34.71 & 35.54 & 35.54 & 35.53 & 35.21 & 35.54 & 34.27 \\
        02-14 & 50 & 225 & 78.58 & 77.70 & 77.70 & 78.17 & 78.17 & 78.16 & 78.17 & 78.17 & 77.10 \\
        02-26 & 24 & 225 & 19.17 & 19.17 & 19.17 & 19.17 & 19.17 & 19.17 & 19.17 & 19.17 & 19.17 \\
        03-14 & 75 & 225 & 72.87 & 63.91 & 64.40 & 71.24 & 71.24 & 70.24 & 69.33 & 71.24 & 67.93 \\
        04-01 & 131 & 225 & 108.30 & 77.07 & 77.88 & 86.81 & 87.33 & 81.57 & 81.22 & 85.13 & 71.39 \\
        04-09 & 61 & 225 & 39.31 & 30.49 & 30.71 & 36.92 & 36.92 & 37.45 & 36.10 & 36.92 & 30.84 \\
        04-24 & 159 & 225 & 127.52 & 95.53 & 102.68 & 114.84 & 114.84 & 115.98 & 112.29 & 113.95 & 107.37 \\
        05-10 & 99 & 225 & 54.40 & 42.01 & 41.66 & 47.81 & 47.81 & 41.88 & 41.15 & 47.59 & 40.70 \\
        05-25 & 166 & 225 & 157.49 & 119.16 & 122.77 & 132.10 & 132.17 & 127.57 & 125.08 & 133.41 & 98.78 \\
        06-04 & 72 & 225 & 40.98 & 31.28 & 35.70 & 37.53 & 37.53 & 36.99 & 36.16 & 37.53 & 34.85 \\
        06-30 & 89 & 225 & 46.49 & 32.62 & 31.73 & 37.27 & 37.27 & 35.44 & 34.04 & 36.17 & 33.46 \\
        07-22 & 855 & 225 & 2105.68 & 1129.43 & 1148.77 & 1125.13 & 1125.13 & 1164.58 & 1143.22 & 1124.33 & 1112.00 \\
        07-24 & 198 & 225 & 109.14 & 73.49 & 77.93 & 86.13 & 86.13 & 86.21 & 82.29 & 86.33 & 78.15 \\
        08-08 & 213 & 225 & 99.77 & 76.42 & 77.42 & 80.09 & 79.93 & 82.00 & 79.56 & 80.95 & 77.39 \\
        08-18 & 42 & 225 & 12.20 & 10.10 & 10.10 & 11.31 & 11.31 & 11.31 & 10.10 & 11.44 & 10.45 \\
        09-04 & 7 & 225 & 2.13 & 1.16 & 1.16 & 1.16 & 1.16 & 1.16 & 1.16 & 1.16 & 1.16 \\
        09-14 & 49 & 225 & 22.29 & 18.38 & 19.45 & 20.53 & 20.53 & 20.21 & 20.21 & 20.53 & 19.72 \\
        10-02 & 118 & 225 & 64.83 & 54.71 & 56.06 & 58.79 & 58.79 & 56.37 & 58.15 & 58.29 & 55.72 \\
        10-17 & 96 & 225 & 1.63 & 0.77 & 0.77 & 0.71 & 0.71 & 0.71 & 0.86 & 0.79 & 0.75 \\
        11-02 & 34 & 225 & 5.00 & 3.97 & 3.96 & 4.24 & 4.24 & 4.17 & 4.03 & 4.24 & 4.05 \\
        11-13 & 179 & 225 & 60.48 & 51.40 & 52.17 & 52.48 & 52.48 & 54.04 & 53.93 & 53.00 & 42.15 \\
        12-01 & 69 & 225 & 38.90 & 36.34 & 36.52 & 37.75 & 37.75 & 37.21 & 37.54 & 37.60 & 35.66 \\
        12-27 & 174 & 225 & 217.59 & 195.65 & 196.20 & 200.64 & 200.64 & 202.56 & 197.48 & 200.90 & 195.40 \\
        \midrule
        \textbf{Total} & 3737 & 5625 & 6126.09 & 4446.02 & 4539.04 & 4591.61 & 4590.66 & \textbf{4655.93} & 4611.51 & 4569.02 & 4321.15 \\
        \textbf{Avg.} & 149.48 & 225 & 245.04 & 177.84 & 181.56 & 183.66 & 183.63 & \textbf{186.24} & 184.46 & 182.76 & 172.85 \\
        \midrule
        \multicolumn{4}{c}{\textbf{Accuracy [\%]}} & 72.58 & 74.09 & 74.95 & 74.94 & \textbf{76.00} & 75.28 & 74.58 & 70.54 \\
        \bottomrule
    \end{tabular}
    \label{sm:aggShed}
\end{table*}

\begin{table*}[htb!]
    \centering
    \small
    \caption{Accuracy of LS detection using selected functional depth metrics with constant number of selected scenarios $n_2= |{\cal O}_k|= 225$ across the 25 test days. $|{\cal E}_k|$ is the number of scenarios with realized LS and GWh refers to the total magnitude of load shedding (across scenarios) on that day. We first screen 550 scenarios based on AUC NL and then using functional depth jointly on grid-wide NL and North Central zonal NL.
    }
    \setlength{\tabcolsep}{2.5pt} 
    \renewcommand{\arraystretch}{1.} 
    \begin{tabular}{c|rrr|rrrrrrrr}
        \toprule
        \textbf{Day} & $|{\cal E}_k|$ & \textbf{$|{\cal O}_k|$} & \textbf{GWh} & \textbf{ID} & \textbf{MBD} & \textbf{EXD} & \textbf{ERD} & \textbf{LID} & \textbf{HMD} & \textbf{DQ} & \textbf{RTD} \\
        \midrule
        01-02 & 681 & 225 & 2573.91 & 2206.82 & 2213.11 & 2175.24 & 2182.11 & 2224.14 & 2241.49 & 2161.78 & 2096.25 \\
        01-20 & 41 & 225 & 31.50 & 29.86 & 30.16 & 31.39 & 31.39 & 31.43 & 31.44 & 31.43 & 31.44 \\
        02-13 & 55 & 225 & 35.92 & 33.41 & 33.41 & 35.35 & 35.35 & 35.20 & 34.65 & 35.37 & 34.06 \\
        02-14 & 50 & 225 & 78.58 & 78.16 & 78.09 & 78.16 & 78.16 & 78.10 & 78.08 & 78.16 & 77.68 \\
        02-26 & 24 & 225 & 19.17 & 19.04 & 19.04 & 19.17 & 19.17 & 19.17 & 19.17 & 19.17 & 19.17 \\
        03-14 & 75 & 225 & 72.87 & 60.16 & 61.31 & 70.62 & 70.62 & 68.04 & 67.19 & 70.62 & 67.07 \\
        04-01 & 131 & 225 & 108.30 & 68.79 & 71.68 & 78.58 & 78.58 & 66.87 & 76.06 & 78.62 & 61.95 \\
        04-09 & 61 & 225 & 39.31 & 26.28 & 27.33 & 34.08 & 34.08 & 35.88 & 32.52 & 34.73 & 28.99 \\
        04-24 & 159 & 225 & 127.52 & 89.66 & 92.88 & 103.99 & 103.99 & 106.98 & 104.67 & 103.39 & 103.26 \\
        05-10 & 99 & 225 & 54.40 & 39.76 & 40.24 & 45.03 & 45.03 & 35.92 & 39.06 & 45.01 & 37.24 \\
        05-25 & 166 & 225 & 157.49 & 105.82 & 108.20 & 120.73 & 122.16 & 112.68 & 113.06 & 122.30 & 85.06 \\
        06-04 & 72 & 225 & 40.98 & 28.37 & 29.71 & 37.34 & 37.34 & 35.16 & 33.96 & 37.34 & 31.26 \\
        06-30 & 89 & 225 & 46.49 & 31.77 & 32.69 & 35.48 & 35.54 & 32.38 & 31.93 & 35.48 & 31.77 \\
        07-22 & 855 & 225 & 2105.68 & 1217.49 & 1249.80 & 1195.93 & 1195.93 & 1303.51 & 1256.67 & 1205.21 & 1244.45 \\
        07-24 & 198 & 225 & 109.14 & 74.61 & 81.52 & 90.24 & 90.24 & 88.59 & 84.60 & 90.07 & 81.87 \\
        08-08 & 213 & 225 & 99.77 & 80.49 & 84.68 & 88.93 & 89.10 & 92.12 & 90.36 & 89.16 & 85.51 \\
        08-18 & 42 & 225 & 12.20 & 9.47 & 9.95 & 11.41 & 11.41 & 9.91 & 10.07 & 11.41 & 9.50 \\
        09-04 & 7 & 225 & 2.13 & 1.16 & 1.16 & 0.84 & 0.84 & 0.84 & 0.84 & 0.84 & 0.84 \\
        09-14 & 49 & 225 & 22.29 & 16.53 & 16.60 & 18.94 & 18.94 & 19.82 & 19.37 & 17.46 & 19.25 \\
        10-02 & 118 & 225 & 64.83 & 50.36 & 54.61 & 57.15 & 57.15 & 55.44 & 55.67 & 56.42 & 53.25 \\
        10-17 & 96 & 225 & 1.63 & 0.78 & 0.50 & 0.69 & 0.71 & 0.71 & 0.63 & 0.71 & 0.74 \\
        11-02 & 34 & 225 & 5.00 & 4.05 & 4.05 & 4.22 & 4.10 & 4.05 & 4.05 & 4.10 & 4.05 \\
        11-13 & 179 & 225 & 60.48 & 50.23 & 50.63 & 51.39 & 51.40 & 51.91 & 51.92 & 51.43 & 50.73 \\
        12-01 & 69 & 225 & 38.90 & 34.13 & 34.47 & 37.52 & 37.60 & 36.63 & 36.63 & 37.56 & 34.46 \\
        12-27 & 174 & 225 & 217.59 & 197.34 & 196.47 & 191.11 & 192.69 & 200.33 & 200.34 & 192.69 & 193.68 \\
        \midrule
        \textbf{Total} & 3737 & 5625 & 6126.09 & 4554.55 & 4622.31 & 4613.55 & 4623.66 & \textbf{4745.82} & 4714.44 & 4610.48 & 4483.55 \\
        \textbf{Avg.} & 149.48 & 225 & 245.04 & 182.18 & 184.89 & 184.54 & 184.95 & \textbf{189.83} & 188.58 & 184.42 & 179.34 \\
        \midrule
        \multicolumn{4}{c}{\textbf{Accuracy [\%]} }  & 74.35 & 75.45 & 75.31 & 75.47 & \textbf{77.47} & 76.96 & 75.26 & 73.19 \\
        \bottomrule
    \end{tabular}
    \label{sm:aggZonalShed_noAdap}
\end{table*}

\begin{figure*}[!htb]
    \centering
    \includegraphics[scale = 0.48, trim = {0cm, 0cm, 0cm, 0cm}, clip]{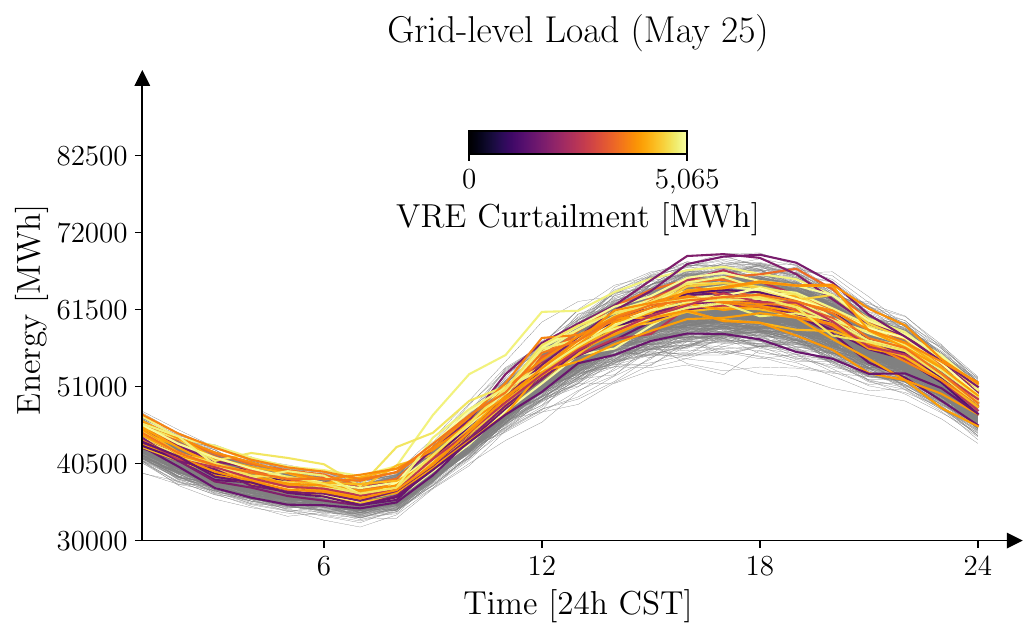}
    \includegraphics[scale = 0.48, trim = {0.9cm, 0cm, 0cm, 0cm}, clip]{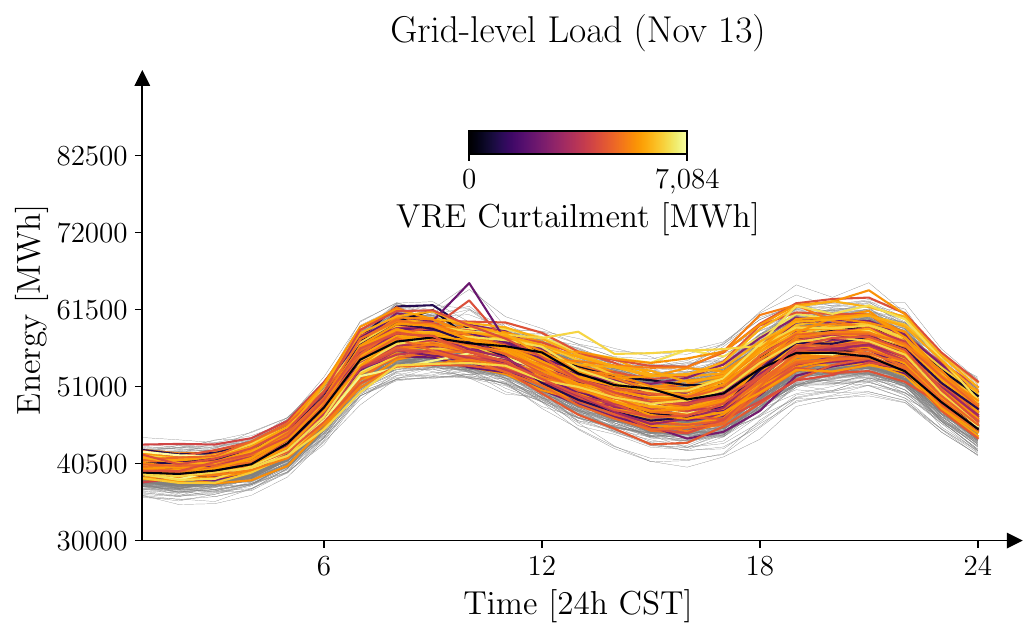}
    \caption{Scenarios with VRE curtailment on May~15 and Nov~13, 2018 illustrating seasonality of load profile. We plot aggregated grid-level load across 24 hours of the day. \emph{Left panel:} winter/spring (cold weather) load profile; \emph{right:}  summer/fall (hot weather) profile. The brighter the color gradient, the more VRE curtailment occurred in the respective scenario.}
    \label{sm:curt_extreme_load}
\end{figure*}

\begin{table*}[htb!]
    \centering
    \small
    \caption{Accuracy of VC detection using selected functional depth metrics with constant number of selected scenarios $n_2 = |{\cal O}_k|= 362$ across the 25 test days. $|{\cal E}_k|$ is the number of scenarios with realized VC and GWh refers to the total variable renewable energy curtailed (across scenarios) on that day. We do not screen scenarios based on AUC and we use functional depth on grid-wide load.}
    \setlength{\tabcolsep}{2.5pt} 
    \renewcommand{\arraystretch}{1.} 
    \begin{tabular}{c|rrr|rrrrrrrrrr}
        \toprule
        \textbf{Day} & $|{\cal E}_k|$ & $|{\cal O}_k|$ & \textbf{GWh} & \textbf{ID} & \textbf{MBD} & \textbf{EXD} & \textbf{ERD} & \textbf{LID} & \textbf{HMD} & \textbf{DQ} & \textbf{RTD} \\
        \midrule
        01-02 & 501 & 362 & 31.53 & 11.77 & 12.30 & 12.39 & 12.42 & 13.15 & 13.07 & 12.70 & 13.77 \\
        01-20 & 578 & 362 & 321.85 & 118.62 & 116.67 & 126.88 & 126.88 & 106.09 & 102.87 & 129.44 & 116.14 \\
        02-13 & 115 & 362 & 76.97 & 39.57 & 41.50 & 55.42 & 55.42 & 50.83 & 56.05 & 57.71 & 50.39 \\
        02-14 & 801 & 362 & 773.98 & 263.74 & 265.72 & 261.25 & 261.72 & 239.88 & 227.02 & 269.43 & 265.00 \\
        02-26 & 318 & 362 & 7.35 & 2.40 & 2.22 & 2.29 & 2.29 & 2.24 & 2.23 & 2.28 & 2.66 \\
        03-14 & 129 & 362 & 37.93 & 19.03 & 18.81 & 19.05 & 19.10 & 19.79 & 18.03 & 20.56 & 18.79 \\
        04-01 & 79 & 362 & 22.59 & 10.38 & 10.43 & 9.68 & 9.68 & 9.33 & 9.03 & 10.35 & 8.47 \\
        04-09 & 36 & 362 & 13.69 & 4.62 & 4.62 & 6.76 & 6.76 & 5.42 & 5.42 & 6.76 & 6.76 \\
        04-24 & 191 & 362 & 39.89 & 17.63 & 17.83 & 15.47 & 15.47 & 15.59 & 16.19 & 15.61 & 18.82 \\
        05-10 & 368 & 362 & 126.12 & 52.15 & 52.20 & 55.97 & 55.91 & 59.36 & 52.93 & 55.27 & 51.76 \\
        05-25 & 40 & 362 & 54.30 & 18.59 & 19.21 & 46.25 & 46.25 & 31.01 & 27.13 & 46.01 & 35.55 \\
        06-04 & 22 & 362 & 26.88 & 5.96 & 13.80 & 22.87 & 22.87 & 22.79 & 17.64 & 22.87 & 21.32 \\
        06-30 & 23 & 362 & 37.72 & 13.02 & 13.73 & 33.45 & 33.45 & 26.85 & 18.47 & 33.45 & 20.08 \\
        07-22 & 883 & 362 & 92.26 & 38.25 & 38.21 & 49.02 & 49.02 & 37.21 & 34.98 & 49.03 & 42.60 \\
        07-24 & 400 & 362 & 49.36 & 26.21 & 26.27 & 39.46 & 39.46 & 34.23 & 25.86 & 39.69 & 35.49 \\
        08-08 & 336 & 362 & 28.53 & 19.83 & 19.91 & 24.07 & 24.08 & 20.00 & 19.60 & 24.16 & 17.34 \\
        08-18 & 33 & 362 & 0.62 & 0.45 & 0.48 & 0.40 & 0.40 & 0.46 & 0.45 & 0.38 & 0.44 \\
        09-04 & 0 & 362 & 0.00 & 0.00 & 0.00 & 0.00 & 0.00 & 0.00 & 0.00 & 0.00 & 0.00 \\
        09-14 & 2 & 362 & 0.01 & 0.01 & 0.01 & 0.01 & 0.01 & 0.01 & 0.01 & 0.01 & 0.01 \\
        10-02 & 7 & 362 & 1.21 & 0.55 & 0.43 & 1.00 & 1.00 & 0.43 & 0.43 & 1.00 & 1.00 \\
        10-17 & 122 & 362 & 14.89 & 7.47 & 7.15 & 6.95 & 6.95 & 6.75 & 6.68 & 6.95 & 6.84 \\
        11-02 & 715 & 362 & 295.17 & 108.97 & 107.65 & 107.49 & 107.76 & 107.82 & 109.75 & 107.08 & 104.15 \\
        11-13 & 100 & 362 & 87.32 & 27.92 & 31.10 & 51.70 & 51.70 & 43.16 & 36.31 & 47.15 & 31.13 \\
        12-01 & 115 & 362 & 18.43 & 7.90 & 10.16 & 7.18 & 7.18 & 9.20 & 9.13 & 7.17 & 7.95 \\
        12-27 & 123 & 362 & 22.15 & 9.81 & 10.16 & 8.98 & 9.11 & 7.95 & 8.12 & 9.50 & 9.19 \\
        \midrule
        \textbf{Total} & 6037 & 9050 & 2180.77 & 824.85 & 840.55 & 963.97 & 964.86 & 869.56 & 817.39 & \textbf{974.54} & 885.64 \\
        \textbf{Avg.} & 241.48 & 362 & 87.23 & 32.99 & 33.62 & 38.56 & 38.59 & 34.78 & 32.70 & \textbf{38.98} & 35.43 \\
        \midrule
        \multicolumn{4}{c}{\textbf{Accuracy [\%]}} & 37.82 & 38.54 & 44.20 & 44.24 & 39.87 & 37.48 & \textbf{44.69} & 40.61 \\
        \bottomrule
    \end{tabular}
    \label{sm:aggCurtail}
\end{table*}

\begin{table*}[htb!]
    \centering
    \small
    \caption{Total VC detection by Far West, North Central, South Central, and West load with no filtering, and not using adaptive selection. DQ was the most suitable depth metric detecting 21.54\% of the VC. The DQ method performs best (62.14\%) when implementing the seasonal approach only in the days defined as extreme and filtering to keep 450 scenarios by AUC in summer and spring season.}
    \setlength{\tabcolsep}{2.5pt} 
    \renewcommand{\arraystretch}{1.} 
    \begin{tabular}{c|rrr|rrrrrrrrrr}
    \toprule
    \textbf{Day} & $|{\cal E}_k|$ & $|{\cal O}_k|$ & \textbf{GWh} & \textbf{ID} & \textbf{MBD} & \textbf{EXD} & \textbf{ERD} & \textbf{LID} & \textbf{HMD} & \textbf{DQ} & \textbf{RTD} \\
    \midrule
    01-02 & 68 & 76 & 22.26 & 6.81 & 6.81 & 2.97 & 2.97 & 6.69 & 7.44 & 5.88 & 4.51 \\
    02-13 & 95 & 76 & 76.21 & 3.21 & 3.66 & 14.86 & 14.86 & 23.29 & 20.94 & 24.07 & 14.21 \\
    02-26 & 13 & 76 & 3.12 & 0.00 & 0.00 & 0.16 & 0.16 & 0.37 & 0.18 & 0.00 & 0.18 \\
    03-14 & 87 & 76 & 35.79 & 6.50 & 6.60 & 5.09 & 4.69 & 5.62 & 5.19 & 6.09 & 7.01 \\
    04-01 & 61 & 76 & 21.98 & 2.00 & 1.34 & 2.40 & 2.33 & 2.44 & 3.38 & 3.94 & 2.59 \\
    04-09 & 33 & 76 & 13.61 & 2.67 & 2.67 & 1.84 & 1.84 & 5.41 & 3.25 & 5.98 & 1.52 \\
    04-24 & 118 & 76 & 36.84 & 5.40 & 4.91 & 4.13 & 4.13 & 3.75 & 4.03 & 2.57 & 6.21 \\
    05-25 & 29 & 76 & 54.00 & 39.80 & 39.80 & 49.29 & 49.69 & 52.95 & 52.95 & 52.95 & 50.38 \\
    06-04 & 17 & 76 & 26.55 & 20.18 & 20.18 & 26.55 & 26.55 & 26.55 & 26.55 & 26.55 & 26.55 \\
    06-30 & 19 & 76 & 37.51 & 27.56 & 25.90 & 36.12 & 36.12 & 37.51 & 37.51 & 37.51 & 37.51 \\
    07-22 & 36 & 76 & 38.30 & 30.08 & 30.08 & 33.69 & 33.69 & 36.64 & 36.64 & 36.64 & 36.75 \\
    07-24 & 17 & 76 & 36.64 & 20.39 & 24.64 & 36.16 & 34.78 & 36.64 & 36.64 & 36.64 & 36.64 \\
    08-08 & 15 & 76 & 21.15 & 7.75 & 7.75 & 20.80 & 20.80 & 21.15 & 21.15 & 21.15 & 20.85 \\
    10-17 & 39 & 76 & 9.96 & 5.64 & 5.64 & 6.60 & 6.60 & 6.50 & 6.59 & 6.86 & 5.95 \\
    11-13 & 77 & 76 & 86.69 & 52.51 & 54.21 & 53.55 & 53.55 & 68.26 & 67.65 & 68.88 & 54.22 \\
    12-01 & 49 & 76 & 15.77 & 7.19 & 7.44 & 6.95 & 6.95 & 7.91 & 7.68 & 7.38 & 6.55 \\
    12-27 & 82 & 76 & 20.43 & 2.04 & 2.04 & 1.85 & 1.85 & 1.24 & 1.41 & 2.93 & 2.40 \\
    \midrule
    \textbf{Total} & 855 & 1292 & 556.82 & 239.75 & 243.68 & 303.01 & 301.56 & 342.91 & 339.18 & \textbf{346.01} & 314.02 \\
    \textbf{Avg.} & 50.29 & 76 & 32.75 & 14.10 & 14.33 & 17.82 & 17.74 & 20.17 & 19.95 & \textbf{20.35} & 18.47 \\
    \midrule
    \multicolumn{4}{c}{\textbf{Accuracy [\%]}} & 43.06 & 43.76 & 54.42 & 54.16 & 61.58 & 60.91 & \textbf{62.14} & 56.40 \\
    \bottomrule
    \end{tabular}
    \label{sm:aggZonalCurtail_short}
\end{table*}

}
\end{document}